\documentclass[twoside]{LNGAI21}

\usepackage{LNGAI21macro}

\usepackage{graphicx}
\usepackage{amsmath}
\usepackage{amssymb}

\usepackage{caption}
\usepackage{subcaption}
\usepackage{float}

\newcommand{\AF}{\ensuremath{\mathit{AF}}}
\newcommand{\Lab}{\ensuremath{\mathcal{L}\!\mathit{ab}}}
\newcommand{\In}{\ensuremath{\mathtt{In}}}
\newcommand{\Out}{\ensuremath{\mathtt{Out}}}
\newcommand{\Undec}{\ensuremath{\mathtt{Undec}}}

\newcommand{\atype}{\ensuremath{\text`a}}

\newcommand{\aSet}{\ensuremath{\text`a~\mathtt{Set}}}
\newcommand{\bSet}{\ensuremath{\text`b~\mathtt{Set}}}
\newcommand{\aRel}{\ensuremath{\text`a~\mathtt{Rel}}}

\newcommand{\aLabelling}{\ensuremath{\text`a~\mathtt{Labelling}}}
\newcommand{\att}{{\leadsto}}

\newcommand{\A}{\mathcal{A}}
\newcommand{\B}{\mathcal{B}}

\newcommand{\F}{\mathcal{F}}
\newcommand{\V}{\mathcal{V}}
\newcommand{\T}{\mathcal{T}}

\newcommand{\bo}{o}
\newcommand{\ar}{\Rightarrow}
\newcommand{\limp}{\longrightarrow}
\newcommand{\ldimp}{\longleftrightarrow}
\newcommand{\cmpl}{-}

\def\lastname{Steen and Fuenmayor}

\begin{document}

\begin{frontmatter}

\title{A Formalisation of Abstract Argumentation in 
	Higher-Order Logic }
  \author{Alexander Steen}
  \author{David Fuenmayor}
  \address{\texttt{$\{\texttt{alexander.steen},\texttt{david.fuenmayor}\}$@uni.lu} \\ University of Luxembourg, Department of Computer Science \\ 6, avenue de la Fonte \\ L-4364 Esch-sur-Alzette, Luxembourg}

  \begin{abstract}
    We present an approach for representing
    abstract argumentation frameworks based on an encoding
    into classical higher-order logic.
    This provides a uniform framework for
    computer-assisted assessment of abstract argumentation frameworks
    using interactive and automated reasoning tools.
    This enables the formal analysis and verification of meta-theoretical properties
    as well as the flexible generation of extensions and labellings with respect to
    well-known argumentation semantics.
    
  \end{abstract}


 \end{frontmatter}

\section{Introduction}
Argumentation theory is a relevant and active field of research in artificial
intelligence.
Argumentation frameworks~\cite{DBLP:journals/ai/Dung95} constitute 
the central concept in abstract argumentation. An argumentation framework essentially
is a directed graph in which the nodes of the graph represent abstract arguments
(carrying no further structural properties) and the edges of the graph represent
attacks between arguments. The exact interpretation of the arguments depends on the
application context: Argumentation frameworks have many topical
applications in, among others, non-monotonic reasoning, logic programming and
multi-agent systems~\cite{2018handbook}. 
As an example, in non-monotonic reasoning the abstract arguments may be regarded as
defeasible reasons (or proofs) for certain claims; and the attack relation then formalises
which of them act as counter-arguments against others.

Since the original formulation of Dung in the 1990s, a lot of research has been
conducted concerning algorithmic procedures, complexity aspects, as well as
various extended and related formalisms, cf.~\cite{DBLP:journals/argcom/BaroniTV20}
and references therein.
In this paper, we propose to investigate argumentation frameworks
from the perspective of extensional type theory (ExTT), also
commonly simply referred to as higher-order logic~\cite{sep-type-theory-church}. 
To that end, we present a novel encoding of argumentation
frameworks and their semantics into higher-order logic, enabling the employment
of off-the-shelf automated theorem provers.
We argue that this constitutes a uniform approach to assess argumentation frameworks
under different argumentation semantics, while at the same time enabling computer-assisted
exploration and verification of meta-theoretical properties within the same logical formalism.
We furthermore argue that our approach is flexible in the sense that it allows
the instantiation of the abstract arguments with arbitrary structures, and to compute
acceptable subsets of arguments satisfying complex (higher-order) properties.
Up to the authors' knowledge, there does not exist any other formalisation of
abstract argumentation frameworks within higher-order logic and, in particular,
existing proof assistants. Although there exists related encodings into less
expressive logical formalisms, these cannot allow for both object-level and
meta-level reasoning within the same framework
(cf.\ \S\ref{sec:conclusion} for a more thorough discussion of related work).

In particular, we exemplarily demonstrate how our approach can be used to (a) flexibly
synthesise extensions and labellings for argumentation frameworks
and (b) to conduct (explorative) assessment of meta-theoretical properties of argumentation
frameworks.
The experiments presented in this article were conducted
using the well-established proof assistant Isabelle/HOL~\cite{DBLP:books/sp/NipkowPW02}.
The corresponding Isabelle/HOL source files
for this work are freely available at GitHub~\cite{Sources}. 

This article is an extended version of an earlier version of this work~\cite{LNGAI}. It contains the
following novel contributions: 
\begin{enumerate}
  \item The encoding is generalised, compared to~\cite{LNGAI}, to allow for 
        interactive and automated reasoning with instantiated argumentation frameworks.
        This is done by defining a \emph{relativised} notion of argumentation semantics
        (cf.~\S\ref{sec:encoding}).
  \item Important meta-theoretical properties of the presented encoding are now fully verified
        in Isabelle/HOL for both extension-based semantics and labellings-based semantics
        (cf.~\S\ref{sec:meta-theory}).
  \item The application of interactive proof assistants for meta-theoretical
        reasoning is exemplified by formally assessing the adequacy of the formalised notions within
        the Isabelle/HOL proof assistant, and by applying its integrated automated tools for theory exploration
        (cf.~\S\ref{sec:meta-theory}).
  \item The flexibility of our approach is demonstrated on interactive
        experiments, including the generation of extensions and labellings
        for abstract argumentation frameworks
        and the assessment on the existence of such extensions (cf.~\S\ref{sec:applications}).
\end{enumerate}

The remainder of this article is structured as follows: 
Technical preliminaries about higher-order logic and abstract argumentation are introduced
in \S\ref{sec:preliminaries}. The encoding of abstract argumentation frameworks and their semantics
is presented in \S\ref{sec:encoding}. Subsequently, \S\ref{sec:meta-theory} and \S\ref{sec:applications}
present the meta-theoretical assessment of the presented encoding and its applications, respectively.
Finally, \S\ref{sec:conclusion} concludes and discusses related and further work.

\section{Preliminaries \label{sec:preliminaries}}

 We start this section with a brief exposition of a higher-order logic (HOL) loosely adopted from earlier work of the first author~\cite{steen2018}. Subsequently, we introduce the notion of abstract argumentation frameworks. 
In the present work, the former formalism is employed as an expressive logical language in order to encode the latter.

\subsection{Higher-Order Logic \label{subsec:hol}}
The term \textit{higher-order logic} refers in general to expressive logical formalisms that allow for quantification over predicate and function variables.
In the context of automated reasoning, higher-order logic commonly refers to systems
based on a simply typed $\lambda$-calculus, as originally introduced in the works of Church, Henkin and several others~\cite{DBLP:journals/jsyml/Church40,DBLP:journals/jsyml/Henkin50}.
In the present work, higher-order logic (abbreviated as HOL) is used interchangeably with Henkin's Extensional Type Theory, cf.~\cite[\S 2]{steen2018}, which constitutes the basis of most contemporary higher-order automated reasoning systems.
HOL provides $\lambda$-notation as an expressive binding mechanism to denote unnamed functions, predicates and sets (by their characteristic functions), and it comes with built-in principles of Boolean and functional extensionality as well as type-restricted comprehension (cf.\ further below).

\paragraph{Syntax and Semantics.}
HOL is a typed logic; and all terms of HOL get assigned a fixed and unique type.
The set $\T$ of types is freely generated from a set of
base types $\B\T$ and the function type constructor $\ar$ (written as a right-associative infix operator).\footnote{
  In order to minimise syntactical differences with respect to the formalisation
  in Isabelle/HOL, we use to symbol $\ar$ to denote the function type constructor
  (despite the fact that other different yet similar symbols are often used in the literature).
  For the same reason, $\longrightarrow$ will denote material implication
  throughout the article. This will help to avoid confusion between the different
  (meta-logical) arrow-like symbols used in Isabelle/HOL.
}
Traditionally, the generating set $\B\T$ is taken to include at least two base types, $\B\T \subseteq \{\iota, \bo\}$,
where $\iota$ is interpreted as the type of individuals and $\bo$ as the type of (bivalent) Boolean truth values.

HOL terms of are given by the following abstract syntax
($\tau,\nu \in \T$):
\begin{equation*}
s,t ::= c_\tau \in \Sigma \; | \; X_\tau \in \V \; | \; \left(\lambda X_\tau.\, s_\nu\right)_{\tau\ar\nu}
  \; | \; \left(s_{\tau\ar\nu} \; t_\tau\right)_\nu
\end{equation*}
where $\Sigma$ is a set of constant symbols and $\V$ a set of variable symbols. The different forms of terms above are called \textit{constants}, \textit{variables}, \textit{abstractions} and \textit{applications}, respectively.
We assume that $\Sigma$ contains
equality predicate symbols $=^\tau_{\tau\ar\tau\ar \bo}$ for each
$\tau \in \T$. All remaining logical connectives (including
conjunction $\land_{\bo\ar\bo\ar\bo}$, disjunction $\lor_{\bo\ar\bo\ar\bo}$, 
material implication $\longrightarrow_{\bo\ar\bo\ar\bo}$,
negation $\neg_{\bo\ar\bo}$, universal quantification for predicates
over type $\tau$ denoted $\Pi^\tau_{(\tau\ar\bo)\ar\bo}$) can
be defined as abbreviations using equality and the other syntactical
structures~\cite[\S 2.1]{steen2018}.\footnote{It is worth noting that in HOL there is no strict differentiation between formulas and terms, as in first-order formalisms. Terms of type $\bo$ are customarily referred to as ``formulas''. Analogously, terms of type $\tau\,{\ar}\,\bo$ (for $\tau \in \T$) are suggestively called ``predicates'' (over type $\tau$).}

For simplicity, the binary logical connectives may be written in infix notation,
e.g., the term/formula $p_o \lor q_o$ formally represents the application
$\left(\lor_{\bo\ar\bo\ar\bo} \; p_o \; q_o\right)$.
Also, so-called \emph{binder notation}~\cite{sep-type-theory-church} is used for universal and existential
quantification: The term $\forall X_\tau.\, s_o$ is used as a short-hand for
$\Pi^\tau_{(\tau\ar\bo)\ar\bo} \left(\lambda X_\tau.\, s_o \right)$
and analogously for existential quantification $\exists X_\tau.\, s_o$.
To improve readability, type-subscripts and parentheses are usually omitted
if there is no risk of confusion. 
Note that, by construction, HOL syntax only admits functions that take one parameter;
$n$-ary function applications are represented using \emph{currying}~\cite{sep-type-theory-church}, e.g.,
a first-order-like term such as $f(a,b)$ involving a binary function $f$ and two constants $a$ and
$b$ is represented in HOL
by consecutive applications of the individual constants,
as in $((f_{\iota\ar\iota\ar\iota}\; a_\iota)\; b_\iota)$,
or simply $f\;a\;b$ if omitting parentheses and type subscripts.
Here, the term $(f\; a)$ itself represents a function that is subsequently applied
to the argument $b$.
Also, functional terms may be only \emph{partially applied}, e.g., occurring in terms
like $(g_{(\iota\ar\iota)\ar\iota}\;(f\; a))$, where $f$ is the ``binary'' function from above
and $g_{(\iota\ar\iota)\ar\iota}$ is a higher-order function symbol taking a functional
expression of type $\iota\ar\iota$ as argument.

HOL automation is usually investigated with respect to so-called
\emph{general semantics}, due to Henkin~\cite{DBLP:journals/jsyml/Henkin50},
for which complete proof calculi can be achieved.
Note that standard models for HOL are subsumed by general models
such that every valid formula with respect to general semantics is also
valid in the standard sense.
We omit the formal exposition to HOL semantics at this point and instead
refer to the literature (cf., e.g.,~\cite{steen2018,sep-type-theory-church}
and the references therein). For the remainder of this article,
HOL with general semantics is assumed.

\paragraph{HOL automation.}
Automated theorem proving (ATP) systems are computer programs that, given
a set of assumptions and a conjecture, try to prove that the conjecture is a
logical consequence of the assumptions. This is done completely autonomously,
i.e., without any interaction from the outside by the user.
In contrast, proof assistants -- also called interactive theorem provers --
allow for the computer-assisted creation and assessment of verified formal proofs,
and also facilitate interactive (and possibly incremental)
experiments with such formal representations. In the interactive scenario,
it is the user that will manually construct, formalise and enter the
proofs into the system. These proofs are then assessed for correctness by the system.
Proof assistants are usually based on (extensions of) higher-order formalisms.
Isabelle/HOL~\cite{DBLP:books/sp/NipkowPW02} is a well-established
HOL-based proof assistant that is employed in a wide range of applications, including this work.
Further well-known proof assistants include, e.g., Coq, Lean, HOL4 and HOL-Light.

One of the most relevant practical features of Isabelle/HOL is the 
Sledgehammer system~\cite{DBLP:journals/jar/BlanchetteBP13} that bridges
between the proof assistant and external ATP systems, 
such as the first-order ATP system E~\cite{DBLP:journals/aicom/Schulz02} 
or the higher-order ATP system Leo-III~\cite{SB2021}, and SMT solvers
such as Z3~\cite{DBLP:conf/tacas/MouraB08} and CVC4~\cite{CVC4}. 
The idea is to use these systems to automatically resolve
open proof obligations and to import the resulting proofs into
the verified context of Isabelle/HOL.
The employment of Sledgehammer is of great practical importance and 
usually a large amount of laborious proof engineering work can be solved
by the ATP systems. In fact, most of the formal proofs presented in the
remainder of this article were automatically constructed using Sledgehammer.
Additionally, Isabelle/HOL integrates so-called \emph{model finders}
such as Nitpick~\cite{Nitpick} that can generate (counter-)models to given formulas.
Also, most non-theorems presented in this work were automatically refuted by Nitpick.

\subsection{Abstract Argumentation \label{ssec:prelim:argumentation}}
The subsequent brief introduction of abstract argumentation frameworks
largely follows the survey paper by Baroni, Caminada and Giacomin~\cite{BCG11} with occasional
references to Dung's seminal paper~\cite{DBLP:journals/ai/Dung95}.
In the present treatment we will not, however, presuppose finiteness for classes of arguments.
Any required cardinality assumptions for argumentation frameworks will be stated explicitly when necessary.
We refer the interested reader to \cite{baumann2017study} (and references therein) for further details on infinite argumentation frameworks.

In the theory of abstract argumentation of Dung~\cite{DBLP:journals/ai/Dung95}, arguments are represented as abstract objects and constitute the nodes of a directed graph.

\begin{definition} 
	An \emph{argumentation framework} $\AF$ is a pair $\AF = \left(\A, \att \right)$,
	where $\A$ is a set (finite or infinite) and ${\att} \subseteq A \times A$ is a binary relation on $\A$.
	The elements of $\A$ are called \emph{arguments}, and $\att$ is called the \emph{attack relation}.
\end{definition}

An argumentation framework formally captures how arguments interact (i.e., conflict
with each other).
Given an argumentation framework $\AF$, 
the primary problem is to determine subsets of arguments that can 
reasonably be accepted together; those sets are called \textit{extensions}.
Restrictions on this selection are imposed by so-called
argumentation semantics.
The set of designators (names) for the different argumentation semantics
is denoted $\mathcal{S}em$ in the following; the restrictions they impose
on the set of extensions are then assigned by an interpretation function.

\begin{definition} 
An \emph{extension-based semantics interpretation} $\mathcal{E}$ associates with each
argumentation framework $\AF = \left(\A, \att \right)$ and an argumentation semantics $S \in \mathcal{S}em$
a set of \emph{extensions}, denoted $\mathcal{E}_S(\AF)$, where $\mathcal{E}_S(\AF) \subseteq 2^{\A}$.
\end{definition}
Roughly speaking, each $E \in \mathcal{E}_S(\AF)$ is a subset of arguments that can be accepted
(under the criterion specified by $\mathcal{E}_S$), while all arguments in $\A \setminus E$ are rejected. In fact, we will show in \S\ref{sec:encoding} how to encode the criteria imposed by $\mathcal{E}_S$ as HOL predicates for several well-known semantics in the literature \cite{BCG11}.

Alternatively, argumentation semantics can be specified in terms of labelling functions (this approach can be traced back to~\cite{caminada2006issue}). We loosely follow \cite{BCG11} below:

\begin{definition}\label{def:labelling} 
Let $\AF = \left(\A, \att \right)$ be an argumentation framework.
A \emph{labelling} of $\AF$ is a total function $\Lab: \A \ar \{\In, \Out, \Undec \}$;
the set of all labellings of $\AF$ is denoted $\mathfrak{L}(\AF)$.
A \emph{labelling-based semantics interpretation} $\mathcal{L}$ then associates with each
$\AF$ and argumentation semantics $S \in \mathcal{S}em$ a set of \emph{labellings}, denoted $\mathcal{L}_S(\AF)$,
where $\mathcal{L}_S(\AF) \subseteq \mathfrak{L}(\AF)$.
\end{definition}
Intuitively, the labels $\In$ and $\Out$ represent the status of 
accepting and rejecting a given argument, respectively. Arguments labelled
$\Undec$ are left undecided, either because one explicitly refrains from accepting resp. rejecting it,
or because it cannot be labelled otherwise.
Given a labelling $\Lab$ we write $(\textit{in}~\Lab)$, $(\texttt{out}~\Lab)$, and $(\textit{undec}~\Lab)$
(read as \textit{in-set}, \textit{out-set}, \textit{undec-set}, respectively)
for the subset of arguments labelled by $\Lab$ as $\In$, $\Out$ and $\Undec$, respectively.

In this paper we will work mainly with the different extension-based semantics introduced by Dung,
together with their labelling-based counterparts, following the exposition by \cite{BCG11}.
Accepted sets of arguments under each of these semantics are termed:
\emph{conflict-free}, \emph{admissible}, \emph{complete}, \emph{grounded}, \emph{preferred}
and \emph{stable}~\cite{DBLP:journals/ai/Dung95} extensions.\footnote{
For reasons of uniformity, we refer to all types of `acceptable' argument sets (according to some given criteria) as \textit{extensions}. In particular, this includes conflict-free sets and admissible sets as well, to which we will also refer as \textit{extensions} in the following.
}
We will also consider some further semantics, namely, \emph{semi-stable}~\cite{DBLP:journals/logcom/CaminadaCD12}, \emph{ideal}~\cite{DBLP:journals/ai/DungMT07} and \emph{stage}~\cite{verheij2003deflog} semantics.
For each of these semantics there exists an equivalent labelling-based formulation.
In fact, there is a one-to-one correspondence between the extension-based semantics
and their labelling counterparts for the semantics listed below, so that
for each extension a corresponding labelling can be found and
vice versa~\cite{DBLP:journals/sLogica/CaminadaG09,BCG11}. This is also why
the names of the argumentation semantics stand for both extension-based
and labelling-based approaches, i.e., in the remainder of this article we consider
the set of argumentation semantics $\mathcal{S}$ to be defined as follows:
\begin{align*}
  \mathcal{S}em = \{ & \texttt{conflictfree}, \texttt{admissible}, \texttt{complete}, \texttt{grounded}, \\
               &    \texttt{preferred}, \texttt{stable}, \texttt{semistable}, \texttt{ideal},
                    \texttt{stage} \}
\end{align*}

We omit the formal definitions for the different extension- and labelling-based semantics
at this point, as they will be subject of the discussion in \S\ref{sec:encoding}.

\section{Encoding of Abstract Argumentation in HOL \label{sec:encoding}}

In this section we discuss the encoding of argumentation semantics in HOL (as introduced in \S\ref{subsec:hol}). For the sake of the formal assessment of the encoding and the verification of our results, the proof assistant Isabelle/HOL~\cite{DBLP:books/sp/NipkowPW02} is employed. A few remarks are in order.

In the discussion below, we will often employ type variables that stand for fixed but arbitrary (base or functional) types. Following Isabelle/HOL's notation, type variables will be represented using letters preceded by a single quote, e.g., $\atype$ is a type variable.
Throughout this work we make generous use of definitions as understood in the context of Isabelle/HOL.
A \emph{definition} defines a new symbol that can, for the purposes of this paper, be regarded
as an abbreviation for the respective terms.
We write $c := s$ to denote the introduction of a new symbol
$c$, with definition $s$, where $s$ is some HOL term.
A \emph{type synonym} is similar to a term definition but rather
introduces a new type symbol that abbreviates a (complex) type expression.

We will often mention several results concerning abstract argumentation as having been proven using Isabelle/HOL. By this we mean formal and internally verified proofs that have been automatically generated by different theorem proving systems integrated into the Isabelle proof assistant.\footnote{Isabelle allows for the manual formalisation of proofs using the general purpose proof language \emph{Isar}~\cite{wenzel2007isabelle}. Isabelle also supports the invocation of external \textit{state-of-the-art} theorem provers via \textit{Sledgehammer} \cite{blanchette2016hammering}. These provers can be run on a local installation or remotely via \textit{System on TPTP} (\url{http://www.tptp.org/cgi-bin/SystemOnTPTP}).}

\begin{figure}[tb] 
	\centering
	\includegraphics[width=.7\textwidth,interpolate]{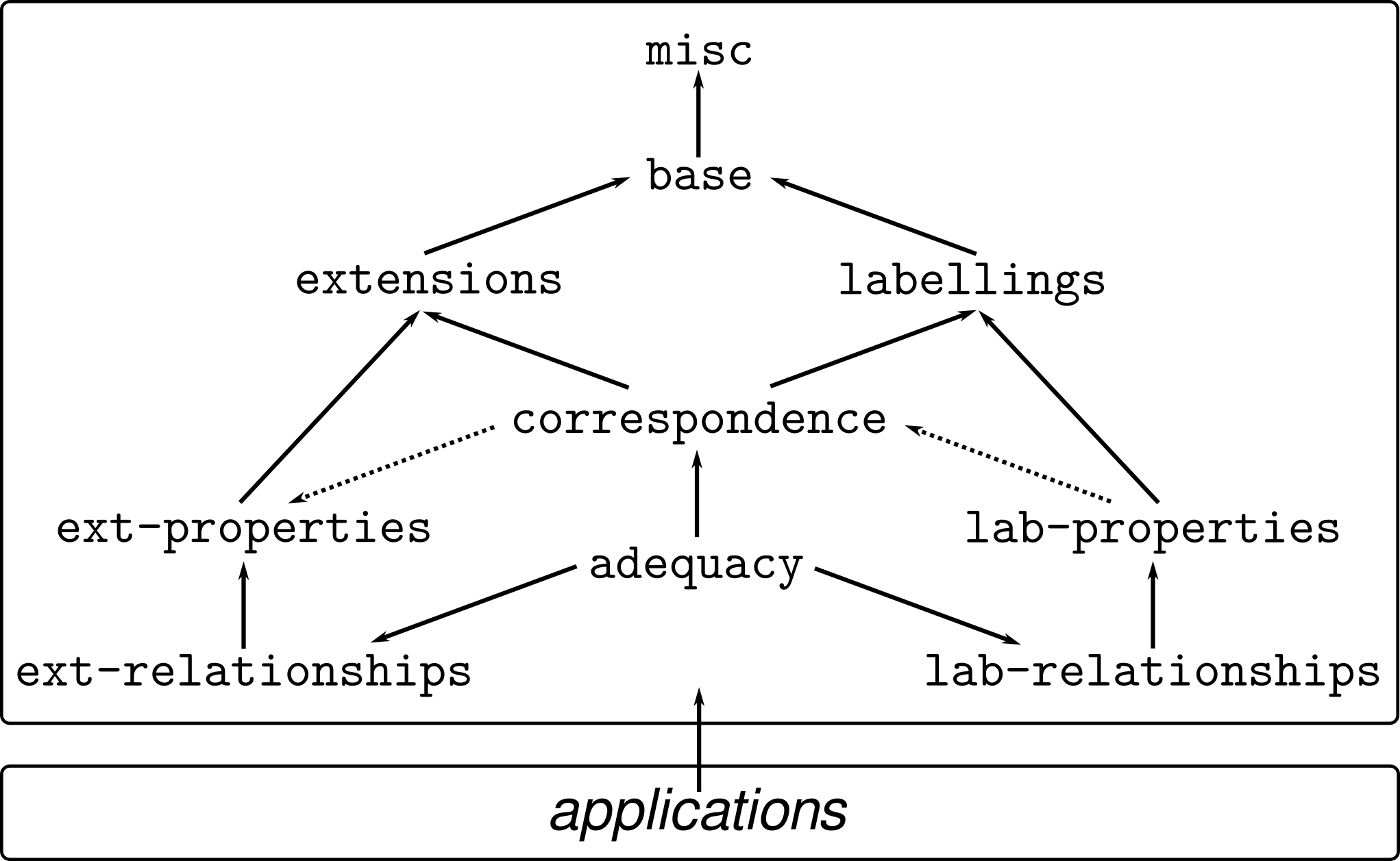}
	\caption{Structure of the encoding as implemented in Isabelle/HOL. The individual nodes
		of the graph represent topically self-contained parts of the overall encoding.
		The solid
		arrows indicate a dependency relation in which the respective part of the encoding
		reuses notions and definitions of the underlying part (e.g., both the definitions
		of the different labellings and extensions use low-level definitions
		collected in the \texttt{base} theory). A dotted arrow indicates auxiliary
		usage where certain lemmas are used as parts of larger proofs.}
		\label{fig:thy:structure}
\end{figure}

The Isabelle/HOL sources for the presented encoding into HOL~\cite{Sources} have been hierarchically organised
into several different files (also referred to as \emph{theory} files).
The general layout is depicted in Fig.~\ref{fig:thy:structure}.
Theory \texttt{misc} collects general purpose notions such as set-theoretic
definitions and notions related to orderings (cf.~\S\ref{subsec:basic}).
Building on this, the theory file \texttt{base} contains general
definitions related to abstract argumentation frameworks (cf.~\S\ref{subsec:basicdefs}).
The different argumentation semantics are then defined in
\texttt{extensions} for extension-based semantics and in 
\texttt{labellings} for labelling-based semantics (cf.~\S\ref{subsec:extension-semantics}
and \S\ref{subsec:labelling-semantics}, respectively).
The remaining theory files contain meta-theoretical results of the presented
encodings, including formal proofs of correspondences between extension-based
and labelling-based semantics, the relationship between different argumentation semantics,
and further fundamental properties. The meta-theoretical assessment results 
are presented separately in \S\ref{sec:meta-theory} and are collected in theory file
\texttt{adequacy}; the remainder of this section
focuses on the encoding in HOL itself.

\subsection{Basic Notions for Sets and Orderings} \label{subsec:basic}

We start by introducing useful type synonyms for the types of sets
and relations, which will be represented by predicates on objects of some type $\atype$.
We thus define $\aSet$ and $\aRel$ as type synonyms for the
functional (predicate) types $\atype\ar \bo$ and $\atype \ar\atype\ar \bo$, respectively.
Set-theoretic operations can be
defined by anonymous functions making use of the respective
underlying HOL logical connectives:
\begin{equation*}
    \begin{gathered}
    \cap \;:=\; \lambda A.\,\lambda B.\,\lambda x.\, (A \; x) \land ( B\;x) 
    \qquad \qquad
    \cup\;:=\; \lambda A.\,\lambda B.\,\lambda x.\, (A\;x) \lor (B\;x) \\
  {\cmpl} \;:=\; \lambda A.\,\lambda x.\, \neg (A\;x)      
    \end{gathered}
\end{equation*}
where $\cap$ and $\cup$ are both terms of type 
$\aSet\ar\aSet\ar\aSet$ and ${\cmpl}$ is of type
$\aSet\ar\aSet$. 
By convention we write binary operations defined this way as infix operators in the remainder, e.g.,
we write $A \cup B$ instead of $(\cup\;A\;B)$.

Set equality
and the subset relation, and also a few other notions defined further below,
will often be used in a \emph{relativised} fashion in the remainder of this article, i.e.,
restricted to a certain subset $D$ of elements of interest.
This is mainly for technical reasons: In HOL, a type $\atype$ intuitively represents 
a set of objects that inhabit the type. Given some type $\atype$ representing
the arguments in our encoding of abstract argumentation, we need to be able
to represent that only a subset of all possible objects of type $\atype$ are 
considered as the domain of arguments (denoted by $\A$; cf.~\S\ref{ssec:prelim:argumentation}) in the argumentation framework. Intuitively,
the extra parameter $D$ (of type $\aSet$) restricts the domain of set-theoretic and logical operations 
to those elements (usually those contained in $\A$). In a dependently typed 
logical formalism as employed, e.g., by the proof assistant Coq, it is possible to encode this information as part of
the type of the operations. In the simply-typed discipline of HOL, however, this needs to be
encoded as part of the term language. This design decision also allows
for the instantiation of the abstract argumentation network with arbitrary (complex)
objects, cf.~\S\ref{subsec:basicdefs} for a more thorough discussion on this topic.\footnote{
In the Isabelle/HOL sources, we also provide a simplified, non-relativised variant 
of these operations. They can be employed when \emph{all objects of the given type} are
considered to be the arguments under consideration. This comes handy in several applications
in abstract argumentation, but not in general (e.g.~when working with instantiated, structured arguments). For reasons of conciseness and legibility, we omit the presentation
of these variants in the article, as they can be obtained seamlessly by simply dropping the superscripts ($D$, resp.~$\A$).
}

Relativised set equality $\approx^D$ and the relativised subset relation $\subseteq^D$,
both of type $\aSet \ar \aSet \ar \aSet \ar \bo$, are
defined by restricting the domain of quantification over the elements in $D$ only, in the following way:
\begin{equation*}\begin{split}
    \approx^{D} \,&:=\, \lambda A.\, \lambda B.\, \forall x.\, (D\;x) \limp \big((A\;x) \ldimp (B\;x)\big) \\
	\subseteq^{D} \,&:=\, \lambda A.\, \lambda B.\, \forall x.\,  (D\;x) \limp \big((A\;x) \limp (B\;x)\big)
\end{split}\end{equation*}

\noindent In a similar spirit, we can define domain-restricted versions of the logical quantifiers, as featured in \textit{free logics}.\footnote{Free logics \cite{sep-logic-free} are quantified, non-classical logics in which terms do not necessarily denote objects in the domain of quantification (sometimes suggestively called ``existing'' objects). Common approaches introduce domain-restricted versions of universal/existential quantifiers, leaving out one or more objects. Free logics have previously been employed to model different notions of partiality in the encoding of axiomatic category theory in Isabelle/HOL \cite{J40}.} Thus, we define the restricted (or relativised) universal quantification predicate $\Pi^D$ as follows:
\begin{equation*}
  \Pi^{D} \,:=\, \lambda P.\, \forall x.\, (D\;x) \limp (P\;x) 
\end{equation*}
We again employ binder notation in the relativised case and write
$\forall^{D} x.\, P$ instead of $\Pi^{D}\;(\lambda X.\, P)$ and let
$\exists^D x.\, P := \neg \big(\forall^D x.\, \neg (P\; x)\big)$.

For the sake of illustration, we note that the above definitions for relativised equality and inclusion can be, alternatively, encoded in a more succinct fashion by employing restricted quantification (we will continue doing so in the sequel).

\begin{equation*}\begin{split}
\approx^{D} \,&=\, \lambda A.\, \lambda B.\, \forall^{D} x.\,(A\;x) \ldimp (B\;x) \\
\subseteq^{D} \,&=\, \lambda A.\, \lambda B.\, \forall^{D} x.\,(A\;x) \limp (B\;x)
\end{split}\end{equation*}

Because of their importance in various argumentation semantics, we
additionally provide generic notions for representing minimal and
maximal (resp.\ least and greatest) sets, with respect to set inclusion:
Let $Obj$ be a term of some type $\atype$, and $Prop$ a predicate of type
$\atype \ar o$. 
We formalise the statement that the set $\varphi(Obj)$ induced by $Obj$ 
is minimal/maximal/least/greatest among all objects $O$
satisfying property $Prop$ wrt.\ a domain of quantification $D$ as follows:
  \begin{equation*}\begin{split}%
    \texttt{minimal}^{D} &:= \lambda Prop.\, \lambda Obj.\, \lambda \varphi.\, (Prop\;Obj) \;\land \\
                         &\qquad \Big(\forall O.\, \big((Prop\;O) \land  (\varphi \;O) \subseteq^{D} (\varphi \; Obj)\big) 
                                  \limp (\varphi\;O) \approx^{D} (\varphi\;Obj)\Big) 
                         \\%
    \texttt{maximal}^{D} &:= \lambda Prop.\, \lambda Obj.\, \lambda \varphi.\, (Prop\;Obj) \;\land \\
                         &\qquad\Big(\forall O.\, \big((Prop\;O) \land  (\varphi \;Obj) \subseteq^{D} (\varphi \; O)\big)
                                  \limp (\varphi\;O) \approx^{D} (\varphi\;Obj) \Big)
                         \\%
    \texttt{least}^{D} &:= \lambda Prop.\, \lambda Obj.\, \lambda \varphi.\, (Prop\;Obj) \;\land \\
                       & \qquad\big(\forall O.\, (Prop\;O) \limp (\varphi\;Obj) \subseteq^{D} (\varphi\;O) \big)  \\%
    \texttt{greatest}^{D} &:= \lambda Prop.\, \lambda Obj.\, \lambda \varphi.\, (Prop\;Obj) \;\land \\
                          &\qquad\big(\forall O.\, (Prop\;O) \limp (\varphi\;O) \subseteq^{D} (\varphi\;Obj) \big)
    \end{split}\end{equation*}%

We formally verified in Isabelle/HOL several well-known properties of least (greatest)
and minimal (maximal) sets while successfully obtaining counter-models for non-theorems
using model finder Nitpick.
As an example, it has been formally verified that least and greatest elements are unique and
that minimal (maximal) elements collapse to the least (greatest) one when the
latter exist~\cite[\texttt{misc}]{Sources}:
\begin{lemma} For every predicate $P$ of type $\aSet \ar o$, elements $O$, $O^\prime$ of type $\atype$
and transformation function $\varphi$ of type $\atype \ar \bSet$ it holds that
\begin{enumerate}\normalfont
  \item $(\texttt{least}^{D}\;P\;O\;\varphi) \land (\texttt{least}^{D}\;P\;O^\prime\;\varphi) \limp (\varphi\;O) \approx^D (\varphi\;O^\prime)$
  \item $(\texttt{greatest}^{D}\;P\;O\;\varphi) \land (\texttt{greatest}^{D}\;P\;O^\prime\;\varphi) \limp (\varphi\;O) \approx^D (\varphi\;O^\prime)$
  \item $(\texttt{least}^{D}\;P\;O\;\varphi) \land (\texttt{minimal}^{D}\;P\;O^\prime\;\varphi) \limp \texttt{least}^{D}\;P\;O^\prime\;\varphi$
  \item $(\texttt{greatest}^{D}\;P\;O\;\varphi) \land (\texttt{maximal}^{D}\;P\;O^\prime\;\varphi) \limp \texttt{least}^{D}\;P\;O^\prime\;\varphi$
\end{enumerate}
\qed
\end{lemma}

Monotonicity of functions over sets is a property that plays an important role in argumentation.
This notion is, again, relativised and encoded as follows:
\begin{equation*}
\texttt{MONO}^{D} := \lambda F.\, \forall A.\, \forall B.\, (A \subseteq^{D} B) \limp (F\;A) \subseteq^{D} (F\;B)
\end{equation*}

We formalised a \textit{fixed point} notion; namely, we speak of sets of arguments being fixed points of operations on sets (e.g.~the so-called \emph{characteristic function} of argumentation frameworks \cite{DBLP:journals/ai/Dung95}).
For a given operation $\varphi$ (of type $\aSet\ar\aSet$) we define in the usual way a fixed-point predicate on sets:
\begin{equation*}
\texttt{fixpoint}^{D} := \lambda \varphi.\, \lambda X.\, (\varphi \; X) \approx^{D} X 
\end{equation*}
We formally verified a weak formulation of the Knaster-Tarski theorem, whereby any monotone function on a powerset lattice has a least (greatest) fixed point~\cite[\texttt{misc}]{Sources}:
\begin{lemma} For every function $f$ of type $\aSet \ar \aSet$ it holds that
\begin{enumerate}\normalfont
  \item $(\texttt{MONO}^{D}\;f) \limp \exists S.\, \texttt{least}^{D}~(\texttt{fixpoint}^{D}\;f)~S~\texttt{id}$
    \item $(\texttt{MONO}^{D}\;f) \limp \exists S.\, \texttt{greatest}^{D}~(\texttt{fixpoint}^{D}\;f)~S~\texttt{id}$
\end{enumerate}
\qed
\end{lemma}

\subsection{Basic Notions for Abstract Argumentation} \label{subsec:basicdefs}

We encode definitions involving some given argumentation framework $\AF = \left(\A, \att \right)$
as HOL terms (i.e., functions) that take as parameters, among others,
the encoded set of arguments $\A$ of type $\aSet$ and the encoded attack relation $att$ of type \aRel.
Note that, for reasons of legibility, the attack relation $att$ will often be referred to as $\att$ in infix notation,
i.e., $a\;\att\;b$ formally stands for $(att\;a\;b)$. 
Sets of arguments are represented by HOL terms of type $\aSet$.
It is worth noting that an abstract argumentation framework $\AF$ can, in principle, be completely
characterised in HOL by simply encoding its underlying attack relation $att$;
this way, the underlying set of arguments $\A$ is given implicitly as the collection of objects of
type \atype\ (i.e., the carrier of $att$)~\cite{LNGAI}. In order to generalise from this specific
case, we do not consider this simplification in the following. This subject is motivated and
discussed at the end of this section.

For a given set of arguments $S \subseteq \A$, we encode its set of attacked ($S^+$) and
attacking ($S^-$) arguments in a relativised fashion as follows:
\begin{equation*}
[\A|att|S]^+ := \lambda b.\, \exists^\A a.\, (S\;a) \land a\;\att\;b 
\qquad
[\A|att|S]^- := \lambda b.\,\exists^\A a.\,(S\;a) \land b\;\att\;a
\end{equation*} 

We now encode the fundamental notion of \textit{defense} of arguments
(called \textit{acceptability} by Dung~\cite{DBLP:journals/ai/Dung95}):
We say that a set of arguments $S$ \textit{defends} an argument $a$ iff
each argument $b \in \A$ attacking $a$ is itself attacked by at least one argument $z$ in $S$.
This is formalised as a HOL predicate \texttt{defends} of type
$\aSet \ar \aRel \ar \aSet \ar \atype \ar o$ by:
\begin{equation*}
\texttt{defends}^{\A} = \lambda att.\, \lambda S.\, \lambda a.\,
             \forall^{\A} b.\, b\;\att\;a \limp (\exists^{\A} z.\, (S\;z) \land z\;\att\;b)
\end{equation*} 
In fact, it can be verified automatically in Isabelle/HOL that the condition imposed
by $\texttt{defends}^{\A} \; att \; S \; a$ can equivalently expressed by $S^+$ and $S^-$
as follows~\cite[\texttt{base}, ll. 53-54]{Sources}:
\begin{lemma}  For every argumentation framework encoded by $(\A, att)$, subset of arguments $S$
and argument $a \in \A$ it holds that
\begin{equation*}\normalfont
\texttt{defends}^{\A} \; att \; S \; a \ldimp [\A|att|\{a\}]^- \subseteq^{\A} [\A|att|S]^+
\end{equation*}
\qed
\end{lemma}

Due to the fact that sets are represented in HOL by predicates (i.e., terms with functional types
returning an element of type $o$),
the notion of \textit{characteristic function} $\F^{\A}$ of an argumentation
framework~\cite{DBLP:journals/ai/Dung95} is actually an extensionally equivalent alias
for the function \texttt{defends$^{\A}$}, yielding:
\begin{equation*}
\F^{\A} := \lambda att.\, \lambda S.\, \lambda a.\, \texttt{defends}^{\A} \; att \; S \; a
\end{equation*}
It is also formally verified that $\F^{\A}$ (i.e., \texttt{defends$^{\A}$}) is indeed a monotone
function and that it has both a least and a greatest fixed point, drawing upon the previously
formalised Knaster-Tarski theorem.

\begin{lemma}
 For every argumentation framework encoded by $(\A, att)$ it holds that
\begin{enumerate}\normalfont
  \item $\texttt{MONO}^{\A}\; (\F^{\A}\;att)$
  \item $\exists S.\, \texttt{least}^{\A}\; \big(\texttt{fixpoint}^{\A}\;(\F^{\A}\;att)\big) \; S \; id$
  \item $\exists S.\, \texttt{greatest}^{\A}\; \big(\texttt{fixpoint}^{\A}\;(\F^{\A}\;att)\big) \; S \; id$
\end{enumerate}
where $id := \lambda x.\; x$ is the identity function.
\qed
\end{lemma}

Recall that argument sets (i.e., potential argument \textit{extensions}) are encoded as functions
mapping objects of an arbitrary type $\atype$ (i.e., arguments) to the bivalent Boolean type $\bo$.
Generalising this, we can now define argument \textit{labellings} as functions into some arbitrary
but finite co-domain of \emph{labels}. 
Following the usual approach in the literature \cite{BCG11}, we assume a fixed set of three labels
$\{\In, \Out, \Undec \}$. This is encoded in HOL as the three-valued type $\texttt{Label}$:\footnote{
For convenience, in our formalisation work \cite{Sources} we have encoded \texttt{Label} as an
Isabelle/HOL datatype, noting that any datatypes can be, in turn, encoded into plain HOL.}
\begin{equation*}
\mathtt{Label}:= \In ~|~ \Out ~|~ \Undec,
\end{equation*}
together with the type synonym $\aLabelling := \atype\ar \mathtt{Label}$ as type of labellings.

It is convenient to encode the basic notions of {in-set}, {out-set} and {undec-set}
of an labelling $\Lab$:
\begin{equation*}\begin{split}
\texttt{in} &:= \lambda \Lab.\,\lambda a.\, (\Lab\;a) = \In \\ 
\texttt{out} &:= \lambda \Lab.\,\lambda a.\, (\Lab\;a) = \Out \\ 
\texttt{undec} &:= \lambda \Lab.\,\lambda a.\, (\Lab\;a) = \Undec
\end{split}\end{equation*}

Using these definitions above we can represent the \textit{as-is-state} of a given argument $a$ wrt.\ a given 
labelling; for instance, $(\texttt{in}\;\Lab\;a)$ means that argument $a$ is labelled \texttt{In} by $\Lab$.

Moreover, we also want to represent a \textit{target-state} in which an argument $a$
is said to be adequately or \textit{legally} labelled. For our particular purposes, we slightly generalise the usual definitions~\cite{BCG11} and say of
an argument $a$ that it is \textit{legally in} if all of its attackers are labelled $\Out$. 
Similarly, $a$ is said to be \textit{legally out} if it has at least one attacker that is labelled $\In$. 
Finally, $a$ is said to be \textit{legally undecided} if it is neither \textit{legally in} nor \textit{legally out}.
These notions are encoded in HOL 
as follows:
\noindent\resizebox{.99\textwidth}{!}{
\begin{minipage}{\textwidth}
\begin{equation*}\begin{split}
\texttt{legallyIn}^{\A} &:= \lambda att.\, \lambda \Lab.\, \lambda a.\,
      \forall^{\A} b.\, (b\;\att\;a) \limp (\texttt{out} \; \Lab)\; b \\
\texttt{legallyOut}^{\A} &:= \lambda att.\, \lambda \Lab.\, \lambda a.\, 
      \exists^{\A} b.\, (b\;\att\;a) \land (\texttt{in}\; \Lab)\; b \\
\texttt{legallyUndec}^{\A} &:= \lambda att.\, \lambda \Lab.\, \lambda a.\,
      \neg(\texttt{legallyIn}^{\A} \; att \; \Lab \; a) \land \neg(\texttt{legallyOut}^{\A} \; att \; \Lab \; a) 
\end{split}\end{equation*}
\end{minipage}
}\medskip

\paragraph{Remark on Relativisation.}
At this point, it might not be apparent what are the benefits of using relativised encodings, 
as opposed to an earlier version of this work in which simpler definitions were
presented~\cite{LNGAI}. 
Recall that HOL is a typed formalism in which each term of the language is associated with a
unique and fixed type. Assuming that abstract arguments are presented by terms of type
$\atype$ in HOL, then the attack relation $att$ is a term of type $\aRel$ which abbreviates the
type $\atype \ar \atype \ar \bo$.
In particular, in~\cite{LNGAI}, argumentation frameworks were completely characterised
by their attack relation; the set of underlying arguments was implicitly assumed 
to be the carrier set of the attack relation, i.e., \textit{all} objects of type $\atype$.

While this seems appealing for assessing properties of argumentation semantics from 
an abstract perspective, the simplified approach is too rigid when concrete arguments
are being studied, e.g., when $\atype$ is instantiated with a type representing formulas
of some logical language. For the sake of the argument, let us assume that $\atype$ is
instantiated with a type $\mathit{PROP}$ of classical propositional logic formulae.\footnote{
Propositional logic can easily be encoded into HOL via so-called \emph{deep embeddings}
in which a new type $\mathit{PROP}$ is postulated, and axiomatised inductively to contain
the respective syntactical elements of propositional logic formulae.
See, e.g., the work by Michaelis and Nipkow on encoding propositional logic
in Isabelle/HOL~\cite{Propositional_Proof_Systems-AFP}.
}
It is clear that the type $\mathit{PROP}$ is generally inhabited by infinitely many 
objects since infinitely many syntactically different propositional logic formulae
can be constructed (assuming a non-empty set of propositional variables). 
This, in turn, implies that every argumentation framework instantiated with $\mathit{PROP}$
and encoded by the simplified representation from~\cite{LNGAI} will be of infinite size,
as there is no possibility to control which objects of type $\mathit{PROP}$ are
contained in the set of arguments $\A$ and which are not.

Clearly this is an undesired effect of implicitly representing the set of arguments as
the carrier of the attack relation.  This is mitigated by the here presented
relativisation in which only a subset of arguments of a certain type are assumed to be
members of the argumentation framework (i.e., those included in the set $\A$).
While relativisation slightly complicates the encoding itself, it allows for a 
more fine-grained control of participating arguments and automatically allows
for instantiating the abstract arguments with arbitrary objects, enabling
the assessment of \emph{instantiated} argumentation frameworks using the very
same encoding. While instantiation is not the primary aim of this work,
we focus on the relativised encoding to allow for future extensions with
instantiated arguments.

\subsection{Extension-based semantics} \label{subsec:extension-semantics}
The well-known extension-based semantics by Dung~\cite{DBLP:journals/ai/Dung95} have been
encoded drawing upon the notions introduced in the previous section.  
For each of the discussed semantics in this section, we first give an informal definition, leaving the underlying argumentation framework $\AF = \left(\A, att \right)$ implicit. We then complement those informal definitions with their corresponding formalisation in HOL, which we encode in their most general form (i.e., relativised wrt.\ the underlying domain or universe of arguments $\A$). 

\subsubsection*{Conflict-free and admissible extensions}

\begin{definition}
	A set of arguments $S$ is termed \textit{conflict-free} if it does not contain two arguments that attack each other.
\end{definition}

\begin{definition}
	A set of arguments $S$ is termed \textit{admissible} if it is conflict-free and defends each of its arguments.
\end{definition}

\paragraph{Formalisation.} Employing the notions encoded in \S\ref{subsec:basicdefs}, we can formalise
the interpretations $\mathcal{E}_{\texttt{conflictfree}}$ and $\mathcal{E}_{\texttt{admissible}}$
as HOL predicates \texttt{conflictfreeExt} and \texttt{admissibleExt}, respectively, of type $\aSet \ar \aRel \ar \aSet \ar \bo$:
\begin{align*}
\texttt{conflictfreeExt}^{\A}:=~& 
\lambda att.\,\lambda S.\;\forall^{\A} a.\,\forall^{\A} b.\,(S \; a) \land (S \; b) \limp \lnot(a~\att~b) \\ 
\texttt{admissibleExt}^{\A}:=~&
\lambda att.\,\lambda S.\; (\texttt{conflictfreeExt}^{\A}~att~S)~\land \\&
\forall^{\A} a.\,(S \; a) \limp (\texttt{defends}^{\A}~att~S~a).
\end{align*}

\subsubsection*{Complete extensions} 

\begin{definition}
An set of arguments $S$ is called a \textit{complete extension} if it is admissible and contains each argument defended by it.
\end{definition}

\paragraph{Formalisation.} The above definition is analogously encoded in HOL:
\begin{align*}
\texttt{completeExt}^{\A}:=~& 
\lambda att.\,\lambda S.\; (\texttt{admissibleExt}^{\A}~att~S)~\land \\&
\forall^{\A} a.~(\texttt{defends}^{\A}~att~S~a) \limp (S~a).
\end{align*}

\subsubsection*{Preferred and grounded extensions}

We now discuss complete extensions which are maximal or minimal wrt.~set inclusion. They are termed \textit{preferred} and \textit{grounded} extensions respectively.

\begin{definition}
	A set of arguments $S$ is termed a \textit{preferred extension} if it is a maximal (wrt.~set inclusion) complete extension.
\end{definition}

\begin{definition}
	A set of arguments $S$ is termed a \textit{grounded extension} if it is a minimal (wrt.~set inclusion) complete extension.
\end{definition}

\paragraph{Formalisation.} The above definitions are encoded as HOL predicates in an analogous fashion:
\begin{align*}
\texttt{preferredExt}^{\A}:=~& 
\lambda att.\,\lambda S.\; (\texttt{maximal}^{\A}~(\texttt{completeExt}^{\A}~att)~S~id) \\
\texttt{groundedExt}^{\A}:=~& 
\lambda att.\,\lambda S.\; (\texttt{minimal}^{\A}~(\texttt{completeExt}^{\A}~att)~S~id)
\end{align*}
Recalling the definitions of maximality (minimality) from~\S\ref{subsec:basic}, these unfold to:
\begin{align*}
\texttt{preferredExt}^{\A}~att~S ~=~& \texttt{completeExt}^{\A}~att~S~\land \\(\forall E.~ \texttt{completeExt}^{\A}&~att~E \land S \subseteq^{\A} E \limp E \approx^{\A} S) \\ 
\texttt{groundedExt}^{\A}~att~S ~=~& \texttt{completeExt}^{\A}~att~S~\land \\ (\forall E.~ \texttt{completeExt}^{\A}&~att~E \land E \subseteq^{\A} S \limp E \approx^{\A} S).
\end{align*}

\subsubsection*{Ideal extensions}\label{subsubsec:ideal-extension}
We will be concerned with admissible sets of arguments that are contained in every preferred extension, which we call \textit{ideal sets}. In any \AF\ the family of ideal sets has indeed an unique maximal (hence greatest) element, which is termed the \textit{ideal extension} \cite{DBLP:journals/ai/DungMT07}.
\begin{definition}
	An \textit{ideal set} is an admissible set of arguments that is a subset of every preferred extension. The (unique) maximal/greatest \textit{ideal set} is called the \textit{ideal extension}.
\end{definition}

\paragraph{Formalisation.} The above are encoded, analogously, as HOL predicates:
		\begin{align*}
		\texttt{idealSet}^{\A}:=~& 
		\lambda att.\,\lambda S.\; (\texttt{admissibleExt}^{\A}~att~S)~\land \\
		&\forall E .~(\texttt{preferredExt}^{\A}~att~E) \limp S \subseteq^{\A} E \\
		\texttt{idealExt}^{\A}:=~& 
		\lambda att.\,\lambda S.\;
		(\texttt{greatest}^{\A}~ (\texttt{idealSet}^{\A}~att)~S~id)
		\end{align*}

\subsubsection*{Stable, semi-stable and stage extensions}\label{subsubsec:sta-extension}

For convenience of exposition we define the \textit{range} of a set of arguments $S$ as the union of $S$ with the set $S^+$ of its attacked arguments. We now turn to sets of arguments whose range satisfies particular maximality requirements.

\begin{definition}
	A set of arguments $S$ is termed a \textit{stable extension} if it is conflict-free and its range is the whole universe (i.e., every possible argument belongs either to $S$ or to $S^+$).
\end{definition}

Stable extensions are in fact complete. However, in contrast to the previous extensions, they do not always exist \cite{DBLP:journals/ai/Dung95}. Semi-stable extensions were introduced independently
by Verheij~\cite{verheij1996two} and Caminada~\cite{DBLP:conf/comma/Caminada06} as an approximate, existence-entailing notion.

\begin{definition}
	A set of arguments $S$ is termed a \textit{semi-stable extension} if it is a complete extension and its range is maximal among all complete extensions.
\end{definition}

While the notion of semi-stable extensions aims at maximising the \textit{range} under the condition of admissibility, stage extensions do so under the (weaker) condition of conflict-freeness.

\begin{definition}
	A set of arguments $S$ is termed a \textit{stage extension} if it is conflict-free and its range is maximal among all conflict-free sets of arguments.
\end{definition}

\paragraph{Formalisation.} The definition of \textit{range} is encoded in HOL as a function of type $\aSet \ar \aRel \ar \aSet \ar \aSet$, while the extension predicates are encoded analogously as before (terms of type $\aSet \ar \aRel \ar \aSet \ar \bo$):
\begin{align*}
\texttt{range}^{\A}:=~& 
\lambda att.\,\lambda S.\;S\cup [\A|att|S]^+ \\
\texttt{stableExt}^{\A}:=~& 
\lambda att.\,\lambda S.\;(\texttt{conflictfreeExt}^{\A}~att~S)~\land~\A\subseteq (\texttt{range}^{\A}~att~S) \\
\texttt{semistableExt}^{\A}:=~& 
\lambda att.\,\lambda S.~ \texttt{maximal}^{\A}~(\texttt{completeExt}^{\A}~att)~S~(\texttt{range}^{\A}~att)\\
\texttt{stageExt}^{\A}:=~& 
\lambda att.\,\lambda S.~ \texttt{maximal}^{\A}~(\texttt{conflictfreeExt}^{\A}~att)~S~(\texttt{range}^{\A}~att)
\end{align*}

\subsection{Labelling-based semantics} \label{subsec:labelling-semantics}
Analogous to the previous section, we provide informal definitions 
followed by their corresponding formalisation in HOL, relativised wrt.~the underlying domain or universe $\A$.

\subsubsection*{Conflict-free and admissible labellings}

\begin{definition}
A labelling $\Lab$ is termed \textit{conflict-free} if (i) every \In-labelled argument is not \textit{legally out}; and (ii) every \Out-labelled argument is \textit{legally out}.
\end{definition}

\begin{definition}
	A labelling $\Lab$ is termed \textit{admissible} if (i) every \In-labelled argument is \textit{legally in}; and (ii) every \Out-labelled argument is \textit{legally out}.
\end{definition}

\paragraph{Formalisation.} Employing the notions encoded in \S\ref{subsec:basicdefs} for argument labellings,  we can formalise
the interpretations $\mathcal{L}_{\texttt{conflictfree}}$ and $\mathcal{L}_{\texttt{admissible}}$
as HOL predicates \texttt{conflictfreeLab} and \texttt{admissibleLab}, respectively, of type $\aSet \ar \aRel \ar \aLabelling \ar \bo$):
\begin{align*}
\texttt{conflictfreeLab}^{\A} :=~ \lambda att .~\lambda \Lab.~ \forall^{\A} &x.~(\texttt{in}~\Lab~x \limp \neg\texttt{legallyOut}^{\A}~att~\Lab~x) \\ &\land\,(\texttt{out}~\Lab~x \limp \texttt{legallyOut}^{\A}~att~\Lab~x) \\
\texttt{admissibleLab}^{\A}   :=~ \lambda att .~\lambda \Lab.~ \forall^{\A} &x.~(\texttt{in}~\Lab~x \limp \texttt{legallyIn}^{\A}~att~\Lab~x) \\ &\land\,
 (\texttt{out}~\Lab~x \limp \texttt{legallyOut}^{\A}~att~\Lab~x)
\end{align*}

We have, in fact, employed Isabelle to verify automatically that admissible labellings always exist (e.g., consider a labelling where all arguments are \Undec) and also that admissible labellings are indeed conflict-free.

Moreover, it has been proven automatically that, for admissible labellings, if an argument is \textit{legally undec} then it is labelled \Undec, but not the other way round (counter-models
provided by Nitpick). Interestingly, one can also verify, again by generating counter-models with Nitpick, that for admissible labellings, a \textit{legally in} (resp. \textit{legally out}) argument is not generally labelled \In~(resp.~\Out). This situation changes, however, when we start considering complete labellings.

\subsubsection*{Complete labellings}

\begin{definition}
	A labelling $\Lab$ is termed \textit{complete} if (i) it is admissible; and (ii) every \Undec-labelled argument is \textit{legally undec}.
\end{definition}
\paragraph{Formalisation.} The above definition is analogously encoded in HOL:
\begin{align*}
\texttt{completeLab}^{\A} :=~\lambda att .~\lambda \Lab.&~ (\texttt{admissibleLab}^{\A}~att~\Lab)~\land \\ 
\forall^{\A} x.&~(\texttt{undec}~\Lab~x)\,\limp\,(\texttt{legallyUndec}^{\A}\,att\,\Lab\,x)
\end{align*}

Using the Sledgehammer tool integrated into Isabelle/HOL it can be proven automatically that
for complete labellings, \textit{legally in} (resp.~\textit{legally out}) arguments are indeed labelled \In~(resp.~\Out). In fact, the following alternative characterisation for complete labellings has been verified as a theorem:
\begin{lemma}\normalfont
\begin{equation*}\begin{split}
\texttt{completeLab}^{\A}~att~\Lab = \forall^{\A} &x.~(\texttt{in}~\Lab~x \ldimp \texttt{legallyIn}~att~\Lab~x) \\
&\land\,(\texttt{out}~\Lab~x \ldimp \texttt{legallyOut}~att~\Lab~x)
\end{split}\end{equation*}
\qed
\end{lemma}

Additionally, it is verified that for complete labellings, we have that \textit{in/out-sets}
completely determine the labelling, i.e., the following two statements hold:
\begin{lemma}\normalfont
\begin{equation*}\begin{split}
(i) \;\; (\texttt{completeLab}^{\A}~att~L_1) \land (\texttt{completeLab}^{\A}~att~L_2) \limp& \\ \;\;\;\;\;\;
\big((\texttt{in}~L_1) \approx^{\A} (\texttt{in}~L_2) \limp \forall^{\A} &x.~(L_1~x) = (L_2~x)\big)
\\
(ii) \; (\texttt{completeLab}^{\A}~att~L_1) \land (\texttt{completeLab}^{\A}~att~L_2) \limp& \\ \;\;\;\;\;\;
\big((\texttt{out}~L_1) \approx^{\A} (\texttt{out}~L_2) \limp \forall^{\A} &x.~(L_1~x) = (L_2~x)\big).
\end{split}\end{equation*}
\qed
\end{lemma}

By generating counter-examples with Nitpick it is verified that, in contrast, \textit{undec-sets}
do not completely determine the (complete) labellings.

\subsubsection*{Preferred and grounded labellings}

We now turn to the notions of minimality and maximality for complete labellings, drawing upon the definitions provided in \S\ref{subsec:basic}. With these notions we can now discuss complete labellings where  \textit{in-sets} are maximal or minimal. They correspond to the so-called \textit{preferred} and \textit{grounded} labellings, respectively.

\begin{definition}
	A labelling $\Lab$ is termed \textit{preferred} if it is a complete labelling whose \textit{in-set} is maximal (wrt.\ set inclusion) among all the complete labellings.
\end{definition}

\begin{definition}
	A labelling $\Lab$ is termed \textit{grounded} if it is a (in fact: \textit{the}) complete labelling whose \textit{in-set} is minimal (wrt.\ set inclusion) among all the complete labellings.
\end{definition}

\paragraph{Formalisation.} The above definitions are encoded as HOL predicates in an analogous fashion:
\begin{align*}
\texttt{preferredLab}^{\A} :=~\lambda att .~\lambda \Lab.&~ \texttt{maximal}^{\A}~(\texttt{completeLab}^{\A}~att)~\Lab~\texttt{in} \\
\texttt{groundedLab}^{\A} :=~\lambda att .~\lambda \Lab.&~ \texttt{minimal}^{\A}~(\texttt{completeLab}^{\A}~att)~\Lab~\texttt{in}
\end{align*}
Recalling the definitions of maximal (resp.~minimal) in \S\ref{subsec:basic}, they unfold into:
\begin{align*}
\texttt{preferredLab}^{\A}~att~\Lab ~=~& \texttt{completeLab}^{\A}~att~\Lab~\land \\\forall L.~ (\texttt{completeLab}^{\A}~att&~L) \land (\texttt{in}~\Lab) \subseteq^{\A} (\texttt{in}~L) \limp (\texttt{in}~L) \approx^{\A} (\texttt{in}~\Lab) \\ 
\texttt{groundedLab}^{\A}~att~\Lab ~=~& \texttt{completeLab}^{\A}~att~\Lab~\land \\ \forall L.~ (\texttt{completeLab}^{\A}~att&~L) \land (\texttt{in}~L) \subseteq^{\A} (\texttt{in}~\Lab) \limp (\texttt{in}~L) \approx^{\A} (\texttt{in}~\Lab).
\end{align*}

\subsubsection*{Ideal labellings}
The notion of \textit{ideal sets} and their maximal/greatest element (\textit{ideal extension}) from \S\ref{subsubsec:ideal-extension} can analogously be lifted to labellings (cf.~\cite{caminada2011judgment}). In order to do this, an ordering relation on labellings needs to be introduced first:

\begin{definition}
	Let $L_1$ and $L_2$ be two labellings. We say that $L_1$ is less or equally committed than $L_2$ if both the \textit{in-set} resp.~\textit{out-set} of $L_1$ are contained in the \textit{in-set} resp.~\textit{out-set} of $L_2$. We use the notation $L_1 \sqsubseteq L_2$.
\end{definition}

We now employ the definition above to lift the corresponding definition for ideal extensions from \S\ref{subsubsec:ideal-extension}: (i) \textit{ideal sets} become \textit{quasi-ideal labellings}, and (ii) \textit{ideal extensions} (greatest ideal sets wrt.~$\subseteq$) become \textit{ideal labellings} (greatest quasi-ideal labellings wrt.~$\sqsubseteq$). Let us now make this definition official:
\begin{definition}
	A labelling is termed \textit{quasi-ideal} if it is admissible and is less or equally committed than every preferred labelling. The `most committed' among all quasi-ideal labellings (i.e., greatest wrt.\ $\sqsubseteq$) is called the \textit{ideal labelling}.
\end{definition}

\paragraph{Formalisation.} The above definitions are encoded, in an analogous manner, as HOL predicates  (of type $\aSet \ar \aRel \ar \aLabelling \ar \bo$):
		\begin{align*}
		\sqsubseteq^{\A} \,:=\,~~\lambda L_1 .~\lambda L_2.&~(\texttt{in}~L_1 \subseteq^{\A} \texttt{in}~ L_2)~\land~(\texttt{out}~L_1 \subseteq^{\A} \texttt{out}~L_2) \\
		\texttt{qidealLab}^{\A} :=~\lambda att .~\lambda \Lab.&~ (\texttt{admissibleLab}^{\A}~att~\Lab)~\land \\
		&\forall L .~ (\texttt{preferredLab}^{\A}~att~L) \limp \Lab \sqsubseteq^{\A} L \\
		\texttt{idealLab}^{\A} :=~\lambda att .~\lambda \Lab.&~(\texttt{qidealLab}^{\A}~att~\Lab)~\land \\
		&\forall L .~ (\texttt{qidealLab}^{\A}~att~L) \limp L \sqsubseteq^{\A} \Lab
		\end{align*}

\subsubsection*{Stable, semi-stable and stage labellings}

We now turn to those labellings whose \textit{undec-sets} must satisfy some particular minimality requirements. Observe that, for their \textit{in-sets}, this works analogously to the corresponding conditions in \S\ref{subsubsec:sta-extension} involving maximality of their \textit{range}.

\begin{definition}
	A labelling $\Lab$ is termed \textit{stable} if it is a complete labelling whose \textit{undec-set} is empty, i.e., no argument is labelled \Undec.
\end{definition}

\begin{definition}
	A labelling $\Lab$ is termed \textit{semi-stable} if it is a complete labelling whose \textit{undec-set} is minimal (wrt.\ set inclusion) among all complete labellings.
\end{definition}

\begin{definition}
	A labelling $\Lab$ is termed \textit{stage} if it is a conflict-free labelling whose \textit{undec-set} is minimal (wrt.\ set inclusion) among all conflict-free labellings.
\end{definition}

\paragraph{Formalisation.} The above definitions are encoded in HOL analogously:
\begin{align*}
\texttt{stableLab}^{\A}:=&~\lambda att .~\lambda \Lab.~
(\texttt{completeLab}^{\A}~att~\Lab) \land \forall x .~ (\Lab~x) \neq \Undec \\
\texttt{semistableLab}^{\A}:=&~\lambda att .~\lambda \Lab.~ \texttt{minimal}^{\A}~(\texttt{completeLab}^{\A}~att)~\Lab~\texttt{undec} \\
\texttt{stageLab}^{\A}:=&~\lambda att .~\lambda \Lab.~ \texttt{minimal}^{\A}~(\texttt{conflictfreeLab}^{\A}~att)~\Lab~\texttt{undec}
\end{align*}

\section{Assessment of Meta-Theoretical Properties \label{sec:meta-theory}}
In this section, meta-theoretic properties of the presented encoding 
are analysed. We proceed by first briefly exemplifying 
the concept of interactive theory exploration in \S\ref{ssec:meta-theory:exploration} as a
powerful application of computer-assisted (meta-theoretical) reasoning.
Here, the utilisation of proof assistants for exploring and synthesising
meta-theoretical concepts and insights in a dialogue-like setting is discussed.
Secondly, in \S\ref{ssec:meta-theory:semantics},
the formal verification of the argumentation semantics' relationships as well as
selected further fundamental properties is presented.
By doing so, we implicitly present adequacy claims of the encoding of abstract argumentation
into HOL.

\subsection{Interactive Theory Exploration \label{ssec:meta-theory:exploration}}
Consider the following lemma, called \emph{Fundamental lemma} by Dung~\cite{DBLP:journals/ai/Dung95}:
\begin{lemma}[Fundamental Lemma] \label{lemma:dung:fund}
Let $\AF = \left(\A, \att \right)$ and $S \subseteq \A$ be an admissible set of arguments.
For any $a \in \A$ it holds that if $S$ defends $a$, then $S \cup \{a\}$ is admissible.\footnote{
Note that in the original formulation of Dung~\cite{DBLP:journals/ai/Dung95} the notion of
\emph{defence} was rather referred to as ``acceptability''.}\qed
\end{lemma}

Suppose that we want to formulate a corresponding fundamental lemma for labelling-based
semantics. It is well-known that a labelling $\Lab$ can be translated into a corresponding
extension by taking its set of \In-labelled arguments $in(\Lab)$. Following this intuition,
a naive (and, indeed, wrong) adaption of Lemma~\ref{lemma:dung:fund} for labellings
could amount to the following statement:
\smallskip

\emph{
Let $\AF = \left(\A, \att \right)$ and $\Lab$ be an admissible labelling for $\AF$.
For any $a \in \A$ it holds that if $in(\Lab)$ defends $a$, then $\Lab^\prime$
is an admissible labelling for $\AF$, where $\Lab^\prime(a) = \In$ and 
$\Lab^\prime(x) = \Lab(x)$ for every $x \neq a$.
} 

\smallskip
Intuitively, in this approach the labelling $\Lab$ is extended to a new labelling
$\Lab^\prime$ that is identical to $\Lab$ except that $a$ is now labelled $\In$ by
$\Lab^\prime$.
Based on the encoding from \S\ref{sec:encoding}, this is formalised in HOL as follows:
\begin{equation*}\begin{split}
\texttt{admissibleLab}^{\A}~att~\Lab \land \texttt{defends}~att~(\texttt{in}\; \Lab)~a \limp & \\
   \texttt{admissibleLab}^{\A}~att~(\lambda x.\,\texttt{if } x = a \texttt{ then } & \In \texttt{ else } (\Lab \; x))
\end{split}\end{equation*}
The new labelling $\Lab^\prime$ is thereby given by the anonymous function
$$\big(\lambda x.\,\texttt{if } x = a \texttt{ then } \In \texttt{ else } (\Lab \; x)\big)$$
that corresponds to the following function in a more conventional mathematical notation:
\begin{equation*}
  \Lab^\prime \colon x \mapsto \begin{cases} \In & \text{ if $ x = a$},\\ \Lab(x) & \text{ otherwise} \end{cases}
\end{equation*}

Note that the \texttt{if-then-else} statement is merely a syntactic abbreviation
(so-called \emph{syntactic sugar}) and can adequately be represented itself in HOL.
But since Isabelle/HOL supports this
elegant and concise representation as well, we adopt it in the following.

\begin{figure}[tb]
\centering
  \begin{subfigure}{\textwidth}
    \centering
    \includegraphics[width=.98\textwidth,interpolate]{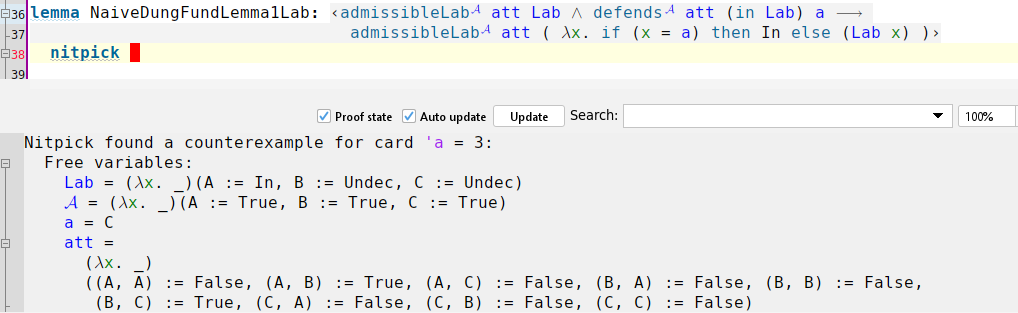}
    \caption{Formalisation of the incorrect fundamental lemma adaption in Isabelle/HOL. 
    The (counter-)model generator \textbf{nitpick} is invoked by writing its name after
    the lemma statement. The generated
    counter-example is printed in the lower part of the window; it encodes a concrete argumentation
    framework and a labelling as human readable (but technical) output.
    Free variables of the statement (in dark blue, left-hand side) are assigned to concrete interpretations
    (in black, right-hand side), e.g., the free variable
    \texttt{A} is assigned to a predicate that maps the synthesised
    arguments \texttt{A}, \texttt{B} and \texttt{C} to true, i.e., representing
    the set $\{\texttt{A}, \texttt{B}, \texttt{C} \}$ of arguments (by its characteristic function).
    Similarly, \texttt{Lab} is assigned to a function that maps
    \texttt{A} to \In, and \texttt{B} and \texttt{C} to \Undec, representing a concrete 
    labelling function $\Lab$.}
    \label{fig:dungfundNitpick1}
  \end{subfigure}
  \bigskip
  \medskip
  
  \begin{subfigure}{\textwidth}
    \centering
    \includegraphics[width=.98\textwidth,interpolate]{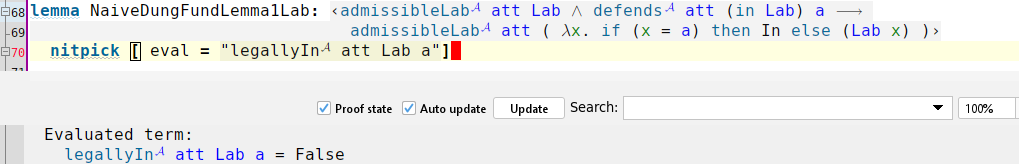}
    \caption{An extended query to \textbf{nitpick} requesting to evaluate (``\texttt{eval}'')
      a concrete expression within the provided counter-model. Here, we ask whether
      the argument assigned to the free variable \texttt{a} is \emph{legally in}; the result
      in the lower part of the windows yields the result that this is not the case
      (the expression is assigned to the value \texttt{False} representing falsehood).}
    \label{fig:dungfundNitpick2}
  \end{subfigure}
  \smallskip
  \caption{Interactive assessment of an exemplary meta-theoretical statement on admissible labellings.}
\end{figure}

When attempting to prove this statement in Isabelle/HOL using the built-in automated
reasoning tools, an invocation of the counter-model generator \textbf{nitpick}
quickly (in less than one second) yields a counter-example to the proposed lemma.
The formulation in Isabelle/HOL and the original output of \textbf{nitpick}
is displayed in Fig.~\ref{fig:dungfundNitpick1}. The false lemma, 
named \texttt{NaiveDungFundLemma1Lab}, is displayed in the upper part of the window;
the lower half contains the output of the counter-model generator.

The counter-example autonomously found by \textbf{nitpick}, cf.\ Fig.~\ref{fig:dungfundNitpick1},
specifies an argumentation framework $\AF = \left(\A, \att \right)$ with
$\A = \{A,B,C\}$, ${\att} = \{(A,B), (B,C) \}$, and an offending initial labelling $\Lab$ given by
$\big(\{A\}, \emptyset, \{B,C\}\big)$.
While this indeed provides a feasible counter-example, it may not be completely apparent why this is the
case. This is why \textbf{nitpick} will also output the offending assignment of any relevant free variable
contained in the statement, here the free variable $\texttt{a}$: In this counter-example,
$\texttt{a}$ is assigned to $C \in \A$;
indicating that the statement is not valid when assuming the above framework $\AF$,
(admissible) labelling $\Lab$ and considering to extend $\Lab$ by additionally labelling $C$
with \In. If it is still not clear why labelling $C$ with $\In$ would result in a labelling
that is not admissible, we can ask \textbf{nitpick} to evaluate further statements in the
context of the found counter-model, e.g., whether $C$ would be \emph{legally in}
in the new labelling $\Lab^\prime$. This is visualised in Fig.~\ref{fig:dungfundNitpick2}.
Of course, any statement can be assessed in this fashion if further information are required
by the user, e.g., in more complex scenarios.

\begin{figure}[tb]
\centering
  \begin{subfigure}{\textwidth}
    \centering
    \includegraphics[width=.48\textwidth,interpolate]{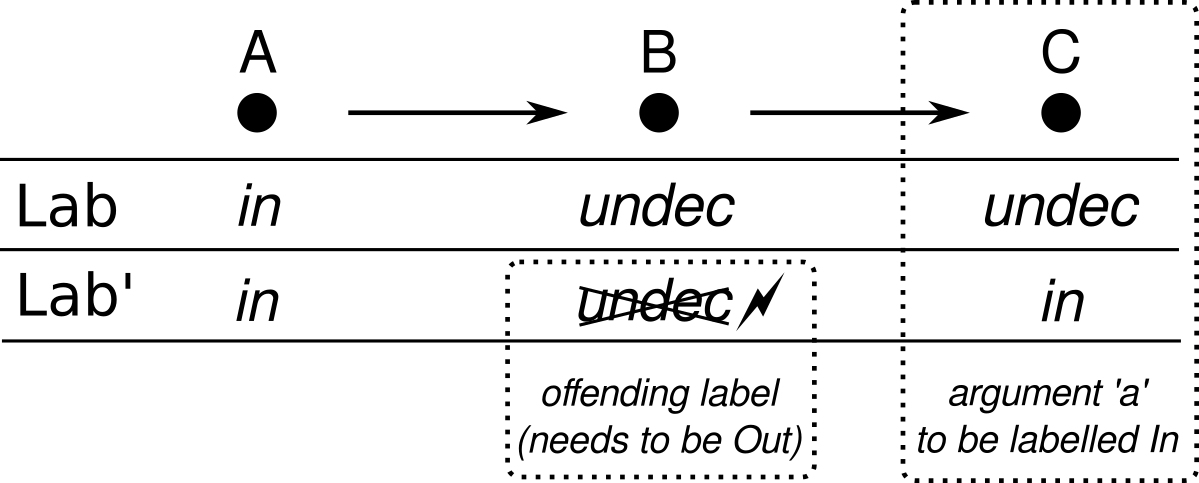}
    \caption{Visualisation of the synthesised counter-example from Fig.~\ref{fig:dungfundNitpick1}.
             Argument $C$ is chosen as the argument to be accepted by the updated labelling, i.e,
             assigned to the free variable $a$ in the lemma formulation.
             The updated labelling $\Lab^\prime$ assigns \In\ to argument $C$ which, in turn,
             requires argument $B$ to be labelled \Out.  }
    \label{fig:dungfundSummary}
  \end{subfigure}
  \bigskip
  \medskip
  
  \begin{subfigure}{\textwidth}
    \centering
    \includegraphics[width=.98\textwidth,interpolate]{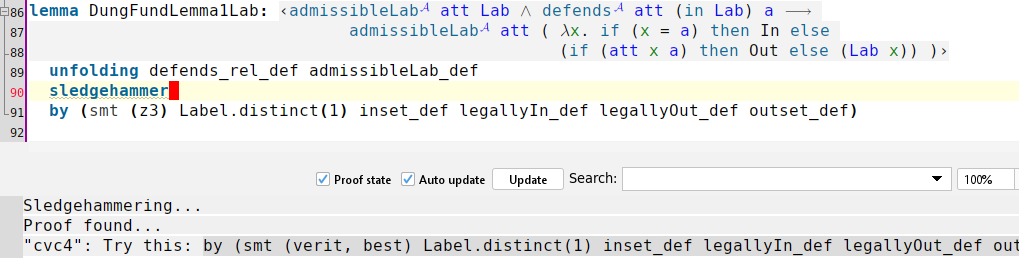}
    \caption{Encoding and verification of the updated adaption of the fundamental lemma
    for admissible labellings. The tool \textbf{sledgehammer} automatically invokes external
    ATP systems and suggests proofs that are subsequently formally verified by Isabelle/HOL.
    Consequently, the one-line proof at line 91 (``\textbf{by} ...'') is a fully
    computer-verified proof of Lemma~\ref{lemma:label:fund}.}
    \label{fig:dungfundLabLemma}
  \end{subfigure}
  \smallskip
  \caption{Analysis of the computer-generated counter-example and formal proof
            of the updated fundamental lemma for labellings.}
\end{figure}

The counter-example, together with the other information provided by the proof assistant, is
summarised in Fig~\ref{fig:dungfundSummary}. Of course, argument $C$ cannot be labelled
\emph{legally in} since, by definition, any attacker of it would need to be labelled out
(but is, in fact, labelled \Undec\ in the counter-example). 
Using this insights, we may propose an updated adapted fundamental lemma for labellings:

\begin{lemma}[Fundamental Lemma for labellings] \label{lemma:label:fund}
Let $\AF = \left(\A, \att \right)$ and $\Lab$ be an admissible labelling for $\AF$.
For any $a \in \A$ it holds that if $in(\Lab)$ defends $a$, then $\Lab^\prime$
is an admissible labelling for $\AF$, where
\begin{enumerate}
  \item $\Lab^\prime(a) = \In$,
  \item $\Lab^\prime(x) = \Out$ if $x~\att~a$, and
  \item $\Lab^\prime(x) = \Lab(x)$ otherwise.
\end{enumerate}\qed
\end{lemma}

\noindent This lemma is translated into HOL syntax as follows:
\begin{align*}
\texttt{admissibleLab}^{\A}~att~\Lab \land \texttt{defends}~att~(\texttt{in}\; \Lab)~a \limp & \\
   \texttt{admissibleLab}^{\A}~att~\big(\lambda x.\,\texttt{if } x = a \texttt{ then } \In &
            \texttt{ else } \\
             (\texttt{if } att~x~a \texttt{ then } & \Out \texttt{ else } (\Lab \; x))\big)
\end{align*}
The updated labelling function $\Lab^\prime$ is again represented by an anonymous function
$(\lambda x.\, \ldots)$ that contains a nested \texttt{if-then-else} statement following
the definition of $\Lab^\prime$ from Lemma~\ref{lemma:label:fund}.

The verification of Lemma~\ref{lemma:label:fund} is displayed at Fig.~\ref{fig:dungfundLabLemma},
where its contents are formalised as lemma \texttt{DungFundLemma1Lab} in Isabelle/HOL.
As depicted in Fig.~\ref{fig:dungfundLabLemma}, the tool \textbf{sledgehammer} may be
used to attempt automated proof search. In this case a proof is found after a few seconds,
provided by the external reasoning system CVC4~\cite{CVC4},
an SMT solver~\cite{DBLP:series/faia/BarrettSST09}, which is
then automatically reconstructed by and verified in Isabelle/HOL.
Hence, we have found a new meta-theoretical property -- a labelling-based fundamental lemma --
using computer-assisted reasoning in an interactive way.

Of course, such an approach may as well be applied to more complex statements in the context of
interactive and semi-automated theory exploration. Note that the encoding of abstract argumentation
into HOL provides a uniform framework for both reasoning \emph{about} abstract argumentation
(i.e., meta-theory) as well as \emph{within} abstract argumentation (i.e., computation of extensions
and labellings) using the same tools. The former case is illustrated in more detail next,
the latter case is addressed in \S\ref{sec:applications}.

\subsection{Adequacy of the Formalisation \label{ssec:meta-theory:semantics}}

One of the key advantages of the presented encoding of abstract argumentation
is that the surrounding logical framework provided by the HOL formalism
allows for the computer-assisted verification of the encoding's adequacy
-- while providing computational means for generating extensions/labellings of
argumentation frameworks at the same time.
Adequacy here means that the semantics of the encoded structures coincide
with the intended ones, e.g., that any extension or labelling produced via the encoding
is sound with respect to the given argumentation semantics. For this purpose, we 
establish the central properties of argumentation frameworks and their semantics for the presented
encoding. The properties are thereby either taken from Dung's original work~\cite{DBLP:journals/ai/Dung95}
or the survey by Baroni et al.~\cite{BCG11}. Each of these properties have been formally
verified within Isabelle/HOL -- usually proven automatically by external
ATP systems in a few seconds using \textbf{sledgehammer}.
This is a strong practical argument for the feasibility of the presented approach
and proof assistants in general.
We present the following properties in their formally encoded variant within HOL,
exactly as given to the proof assistant; free variables are implicitly universally
quantified.
We start by providing general properties of admissible and conflict-free sets represented
by the encoding:

There are no self-attacking arguments in conflict-free extensions~\cite[p. 8]{BCG11}:
\begin{lemma}
{\normalfont
$\texttt{conflictfreeExt}^{\A}\;att\;E \limp \neg\big(\exists^{\A} a.\, (E\;a) \land a\;\att\;a\big)$.
} 
\qed
\end{lemma}
The characteristic function preserves conflict-freeness (monotonicity was already shown
in \S\ref{subsec:basicdefs}):
\begin{lemma} \mbox{} \\
\normalfont
$\texttt{conflictfreeExt}^{\A}\;att\;E \limp \texttt{conflictfreeExt}^{\A}\;att\;(\F^{\A}\;att\;E)$.
\qed
\end{lemma}

A conflict-free set $E$ is admissible iff $E \subseteq \F^{\A}(E)$~\cite[Lemma 18]{DBLP:journals/ai/Dung95}:
\begin{lemma} \mbox{} \\
{\normalfont
$\texttt{conflictfreeExt}^{\A}\;att\;E \limp \big(\texttt{admissibleExt}^{\A}\;att\;E \ldimp E \subseteq^{\A} \F^{\A}\;att\;E\big)$.
}\mbox{}
\qed
\end{lemma}

Admissible sets can be extended with defended arguments (Dung's fundamental lemma)~\cite[Lemma 10]{DBLP:journals/ai/Dung95}:
\begin{lemma}\mbox{}
\begin{enumerate}\normalfont
\item $\left(\texttt{admissibleExt}^{\A} \; att \; E \land \texttt{defends} \; att \; E \; a\right) \\ ~~~~~~~~~~~~~~~~~~~~~~~~~~~~~~~~~~~~~~~~~~~~~~~\limp~
   \texttt{admissibleExt}^{\A} \; att \; (E \cup a)$
\item $\left(\texttt{admissibleExt}^{\A} \; att \; E \land \texttt{defends} \; att \; E \; a \land
  \texttt{defends} \; att \; E \; a^\prime\right) \\
  ~~~~~~~~~~~~~~~~~~~~~~~~~~~~~~~~~~~~~~~~~~~~~~~\limp 
   \texttt{defends}^{\A} \; att \; (E \cup a) \; a^\prime$ 
   \qed
\end{enumerate}
\end{lemma}

Admissible sets form a $(\omega-)$complete partial order (CPO) with respect to set
inclusion~\cite[Theorem 11]{DBLP:journals/ai/Dung95}.
In fact, a stronger statement could be formally verified: The collection of admissible sets form
a directed CPO (dCPO):
\begin{lemma}  
\mbox{}
\begin{enumerate}\normalfont
  \item $\texttt{$\omega$-cpo}^{\A}\;(\texttt{admissibleExt}^{\A}\; att)$
  \item $\texttt{dcpo}^{\A}\;(\texttt{admissibleExt}^{\A}\; att)$ 
\end{enumerate}
where $\texttt{$\omega$-cpo}^{\A}$ and $\texttt{dcpo}^{\A}$ encode the notions
of (directed) complete partial orders in HOL~\cite[\texttt{misc}]{Sources}.
\qed
\end{lemma}

From this, it can be verified that for each admissible set $S$ 
there exists a preferred extensions extending $S$~\cite[Theorem 11]{DBLP:journals/ai/Dung95}:
\begin{lemma}  \mbox{}\\
\normalfont
  $\texttt{admissibleExt}^{\A}\; att \; S
      \limp \exists E.\, S \subseteq^{\A} E \land \texttt{preferredExt}^{\A}\;att\;E$ 
\qed
\end{lemma}

We proceed by highlighting a few properties of selected classes of extensions resp. labellings.
Conflict-free sets are complete iff they are fixed-points of $\F$~\cite[Lemma 24]{DBLP:journals/ai/Dung95}:
\begin{lemma} \mbox{} \\
\normalfont
$\texttt{conflictfreeExt}^{\A}\;att\;S \limp \big(\texttt{completeExt}^{\A}\;att\;S \ldimp S \approx^{\A} \F^{\A}\;att\;S\big)$.
\qed
\end{lemma}

Complete, preferred and grounded extensions always exist~\cite{DBLP:journals/ai/Dung95}:
\begin{lemma} \mbox{}
\begin{enumerate} \normalfont
  \item $\exists E.\, \texttt{completeExt}^{\A}\;att\;E$
  \item $\exists E.\, \texttt{preferredExt}^{\A}\;att\;E$
  \item $\exists E.\, \texttt{groundedExt}^{\A}\;att\;E$
\qed
\end{enumerate}
\end{lemma}

Grounded labellings can equivalently be characterised by minimal in-sets and
minimal out-sets~\cite[Prop. 5]{BCG11}:
\begin{lemma}
\mbox{} \\
\normalfont
$\texttt{minimal}^{\A}\; (\texttt{complete}^{\A}\;att)\;\Lab\;\texttt{in}
 \ldimp \texttt{minimal}^{\A}\; (\texttt{complete}^{\A}\;att)\;\Lab\;\texttt{out}$
\\\mbox{}\qed
\end{lemma}

Furthermore, grounded extensions resp. labellings are unique~\cite[Prop.\ 4]{BCG11}:
\begin{lemma}\mbox{}
\begin{enumerate}\normalfont
  \item $\texttt{groundedExt}^{\A}~att~S \ldimp \texttt{least}^{\A}~(\texttt{completeExt}^{\A}~att)~S~id$
  \item $\texttt{groundedLab}^{\A}~att~\Lab \ldimp \texttt{least}^{\A}~(\texttt{completeLab}^{\A}~att)~\Lab~\texttt{in}$ \qed
\end{enumerate}
\end{lemma}

Analogously, preferred labellings can equivalently be characterised by maximal in-sets and
minimal out-sets~\cite[Prop. 8]{BCG11}:
\begin{lemma}
\mbox{} \\
\normalfont
$\texttt{maximal}^{\A}\; (\texttt{complete}^{\A}\;att)\;\Lab\;\texttt{in}
 \ldimp \texttt{maximal}^{\A}\; (\texttt{complete}^{\A}\;att)\;\Lab\;\texttt{out}$
\\\mbox{}\qed
\end{lemma}

\begin{figure}[tb]
\centering
\includegraphics[width=.85\textwidth,interpolate]{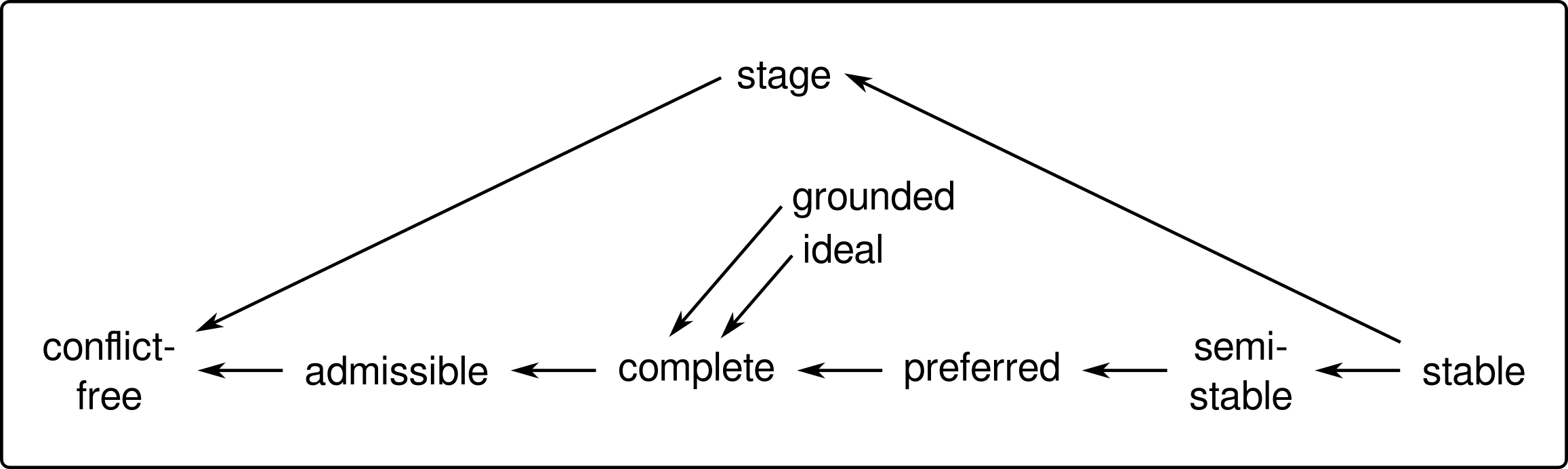}
\caption{Relationship between each of the different argumentation semantics considered in this
article.
Each entry represents both the respective extension-based and labelling-based semantics.
An arrow from one semantics to another can be read as \emph{``is a''}, it symbolises
a generalisation relation, i.e., the latter is more general than the former.
Transitive arrows are omitted.}
\label{fig:semantics}
\end{figure}

Additionally to the properties of the individual argumentation semantics,
there exist well-known relationships between them:
As an example, every complete labelling is also an admissible labelling; and
every stable extension is also a conflict-free one~\cite{BCG11}.
Fig.~\ref{fig:semantics} shows these relationships for a subset of argumentation
semantics considered in this article. 
As another case study for the assessment of meta-theoretical properties of argumentation
semantics using proof assistants,
we have verified all of the displayed links in Isabelle/HOL~\cite{Sources}.

Even more, the non-inclusion of the inverse directions of the arrows from
Fig.~\ref{fig:semantics} has been established, for each case, using the counter-model generator
\textbf{nitpick}. Here, small counter-examples are generated automatically that illustrate
the non-validity of the inverse statement. As an example,
\textbf{nitpick} is able to refute the statement that
every stage labelling is also a stable labelling, encoded in HOL as
\begin{equation*}
 \texttt{stage}^{\A} \; att \; \Lab \limp \texttt{stable}^{\A} \; att \; \Lab,
\end{equation*}
in less than one second.
It is easy to check that the labelling returned by \textbf{nitpick}
is indeed a stage labelling but not a stable one.

Furthermore we formally verified the different correspondence results between
extensions and labellings~\cite{Sources} for the semantics listed below.
\begin{align*}
\mathcal{S}em = \{ & \texttt{conflictfree}, \texttt{admissible}, \texttt{complete}, \texttt{grounded}, \\
&    \texttt{preferred}, \texttt{stable}, \texttt{semistable}, \texttt{stage} \}
\end{align*}

For this sake we have formalised the well known translation mappings between extensions and labellings~\cite{BCG11} as the HOL functions: \texttt{Lab2Ext} and \texttt{Ext2Lab} of types $\aLabelling \ar \aSet$ and $\aSet \ar \aRel \ar \aSet \ar \aLabelling$ respectively. Observe that the function \texttt{Lab2Ext}, in contrast to \texttt{Ext2Lab}, can be defined independently of the underlying argumentation framework.
\begin{align*}
\texttt{Lab2Ext} ~:=~& \lambda \Lab .~ \texttt{in}~\Lab \\
\texttt{Ext2Lab}^{\A} ~:=~& \lambda att .~ \lambda E .~ \lambda a.~ \texttt{if}~ (E~a)~ \texttt{then~In~else} \\
 &\;\;\;\;\texttt{if}~ ([\A|att|E]^+ ~a)~ \texttt{then Out else Undec}
\end{align*}

As an example, the following statements relate preferred extensions and preferred labellings (analogous results hold for the other semantics listed above):

\begin{lemma} \mbox{}
\begin{enumerate}\normalfont
 \item $\texttt{preferredLab}^{\A}\;att\;\Lab \limp  \texttt{preferredExt}^{\A}\;att\;(\texttt{Lab2Ext}\;\Lab)$ \label{item1}
 \item $\texttt{preferredExt}^{\A}\;att\;S \limp \texttt{preferredLab}^{\A}\;att\;(\texttt{Ext2Lab}\;S)$.
 \item $\texttt{preferredLab}^{\A}\;att\;(\texttt{Ext2Lab}\;S) \limp \texttt{preferredExt}^{\A}\;att\;S$.
\qed
\end{enumerate}
\end{lemma}
It is worth noting that the model generator \textit{nitpick} has, in fact, found counter-models to the converse implication to item (\ref{item1}) above, as expected \cite{BCG11}.

In summary, the presented HOL encoding provides not only the definition
of abstract arguments and its different semantics, but also provides
a formal and computer-assisted verification of different meta-theoretical
properties including relationships between argumentation semantics,
correspondences, and equivalent alternative characterisations.

\section{Flexible Generation of Extensions and Labellings} \label{sec:applications}

The encoding of the different argumentation semantics presented above
captures the structure and the logical behaviour of abstract argumentation frameworks within HOL. Building on top of that, we can make
use of \textit{model finders}, e.g., Nitpick~\cite{Nitpick} and Nunchaku~\cite{Nunchaku},
for generating concrete extensions and labellings for a given argumentation frameworks.
To this end we here employ the model finder Nitpick that is readily integrated into the Isabelle/HOL
proof assistant.\footnote{Of course, other automated theorem provers and model generators for higher-order logics, as provided in other proof assistants (e.g.~HOL, Coq, Lean) can also be employed for our purposes. To the best of our knowledge, the level of proof automation featured in the Isabelle/HOL ecosystem is currently unmatched.}

\subsection{Generating Standard Extensions and Labellings}

\begin{figure}[tb]
	\centering
	\begin{subfigure}[b]{0.4\textwidth}
		\centering
		\includegraphics[width=\textwidth]{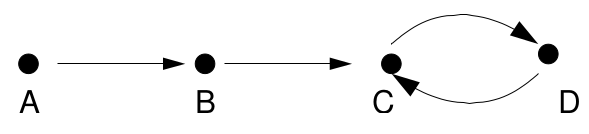}
		\caption{Simple AF}
		\label{fig:examples:4}
	\end{subfigure}
	\hfill
	\begin{subfigure}[b]{0.28\textwidth}
		\centering
		\includegraphics[width=.9\textwidth]{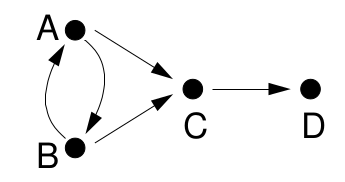}
		\caption{Floating acceptance}
		\label{fig:examples:5}
	\end{subfigure}
	\hfill
	\begin{subfigure}[b]{0.2\textwidth}
		\centering
		\includegraphics[width=\textwidth]{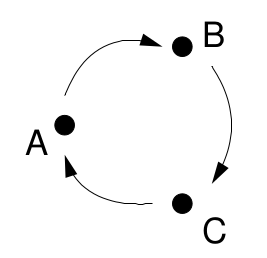}
		\caption{Odd cycle}
		\label{fig:examples:6}
	\end{subfigure}
	\caption{Some representative examples of argumentation frameworks by Baroni, Caminada and Giacomin~\cite[Fig. 4-6]{BCG11}.
		\label{fig:examples}}
\end{figure}

Figure~\ref{fig:examples} displays a few representative examples of argumentation frameworks,
taken from~\cite{BCG11}, that serve as use cases
to illustrate the generation of extensions and labellings employing the model finder Nitpick. We will discuss only a few results. The rest can be consulted in the corresponding Isabelle/HOL sources that can be found
in~\cite[\texttt{model-generation}]{Sources}.

\begin{figure}[htb]
	\centering
	\begin{subfigure}[b]{0.3\textwidth}
		\centering
		\includegraphics[width=\textwidth]{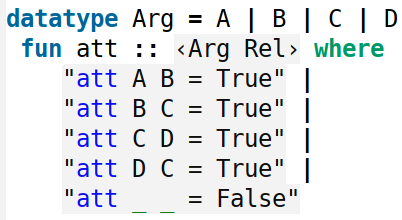}
		\caption{Simple AF}
		\label{fig:examples:4:encoded}
	\end{subfigure}
	\hfill
	\begin{subfigure}[b]{0.3\textwidth}
		\centering
		\includegraphics[width=\textwidth]{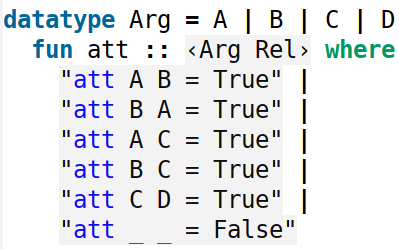}
		\caption{Floating acceptance}
		\label{fig:examples:5:encoded}
	\end{subfigure}
	\hfill
	\begin{subfigure}[b]{0.3\textwidth}
		\centering
		\includegraphics[width=\textwidth]{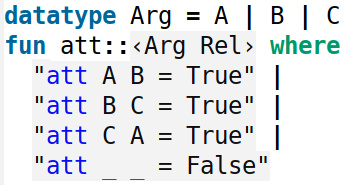}
		\caption{Odd cycle}
		\label{fig:examples:6:encoded}
	\end{subfigure}
	\caption{Encoding the argumentation frameworks in Fig.~\ref{fig:examples}.
		\label{fig:examples:encoded}}
\end{figure}

To encode the above argumentation graphs in Isabelle/HOL we employ the simplified encoding approach as discussed in \S\ref{sec:encoding}. In this approach the universe of arguments is implicitly given as the carrier of the \textit{attack} relation. Thus, for the type of arguments we define a new datatype: \texttt{Arg}, 
consisting only of the distinct terms \texttt{A}, \texttt{B}, \texttt{C} (and
\texttt{D} when required). Next, we encode the attack relation \texttt{att} as binary predicate such that $\texttt{att}~X~Y$ if and only if $X$ attacks $Y$ according to the corresponding graph in Fig.~\ref{fig:examples}. This is  displayed in the corresponding Isabelle/HOL setup in Figure~\ref{fig:examples:encoded}. 

\begin{figure}[tb]
	\fbox{
		\parbox{.95\textwidth}{
			\includegraphics[width=.95\textwidth,interpolate]{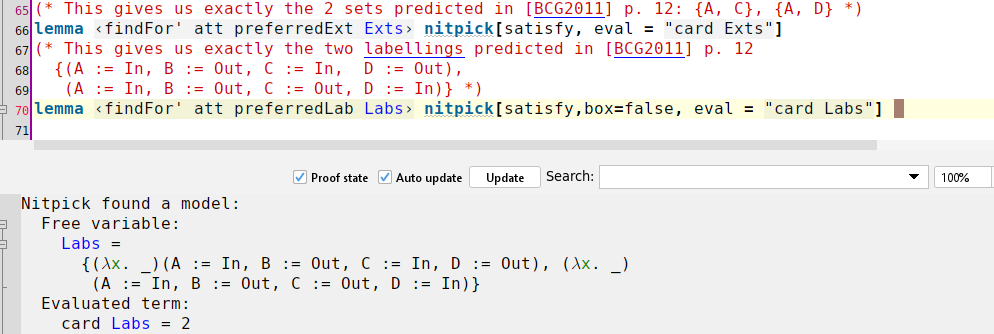}
		}
	}
	\caption{Nitpick output enumerating all preferred labellings of the argumentation
	framework from Fig.~\ref{fig:examples:4}.}
	\label{fig:examples-nitpick-all}
\end{figure}

We can now ask the model finder Nitpick to generate, say, all preferred labellings for the \AF\ in Fig.~\ref{fig:examples:4}. Indeed, Nitpick produces the following two labellings
(cf.\ the original output of Nitpick displayed in Fig.~\ref{fig:examples-nitpick-all}): 
\begin{equation*}
\Lab_1: x \mapsto \, \begin{cases}
\In & \text{if $x = A$}\\
\Out & \text{if $x = B$}\\
\In & \text{if $x = C$}\\
\Out & \text{if $x = D$}\\
\end{cases}
\;\;\;\texttt{and}\;\;\;
\Lab_2: x \mapsto \,
\begin{cases}
\In & \text{if $x = A$}\\
\Out & \text{if $x = B$}\\
\Out & \text{if $x = C$}\\
\In & \text{if $x = D$}\\
\end{cases}
\end{equation*}
which represent the labelling $\Lab_1$ and $\Lab_2$ such that $in(\Lab_1) = \{A,C\}$, $out(\Lab_1) = \{B,D\}$
and $undec(\Lab_1) =\emptyset$, and $in(\Lab_2) = \{A,D\}$, $out(\Lab_2) = \{B,C\}$
and $undec(\Lab_2) =\emptyset$. 

In the example above, we employed a specially engineered function \texttt{findFor}, given by
$$\texttt{findFor} \;:=\; \lambda AF.~\lambda Prop.~\lambda S.~\forall Lab .~(S~Lab) \ldimp (Prop~AF~Lab), $$
in such a way that Nitpick tries to satisfy the statement
 $$\texttt{findFor}~att~\texttt{preferredLab}~Labs$$
by generating a model (and, hence, enumerating all the labellings).
Nitpick provides all preferred labellings by finding the value given to the free variable \textit{Labs} above.
This is equivalent to finding the value of \textit{Labs} such that 
\begin{equation*}
\forall Lab .~(Labs~Lab)~\ldimp~(\texttt{preferredLab}~att~Lab) 
\end{equation*}
holds, where \texttt{preferredLab} is the predicate as defined in \S\ref{subsec:labelling-semantics}.
We observe that the reported results are in fact as described in~\cite{BCG11}. The same holds for the remaining argumentation
semantics and examples from Fig.~\ref{fig:examples}.

\subsection{Generating Flexibly-Constrained Extensions and Labellings}

In addition to the above -- quite standard -- applications, we can now make use of the
expressive surrounding logical framework to ask for \emph{specific} labellings, flexibly constrained by means of an arbitrary (higher-order) predicate. Consider the example
displayed in Fig.~\ref{fig:examples-nitpick-higherorder} extending the example from Fig.~\ref{fig:examples:5}:

\begin{figure}[tb]
	\noindent\fbox{
	\includegraphics[width=.93\textwidth]{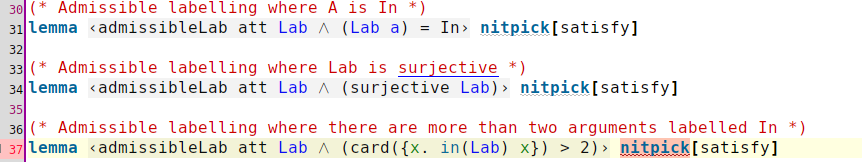}
}
	\caption{Asking for a specific labelling that satisfies additional properties in Isabelle/HOL.}
	\label{fig:examples-nitpick-higherorder}
\end{figure}

In the three lemmas, we ask Nitpick to generate admissible labellings where, additionally,
(1) argument $A$ is labelled \In, (2) $Lab$ is a surjective function, and (3)
there are more than two arguments labelled \In, respectively. In the first two cases, 
suitable labellings are provided, in the third case no such labelling can be found
(visualised by the red background colour indicating an error). Indeed, no such labelling
exists. 

Similarly, we can prove in Isabelle/HOL that for Fig.~\ref{fig:examples:6} no admissible
labelling other than the trivial one exists. 
This is expressed by the formula 
\begin{equation*}
(\texttt{admissibleLab}\;att\;\Lab) \longrightarrow \forall x.\, (\Lab\;x) = \Undec
\end{equation*}
which is proven automatically by Sledgehammer within a few seconds.
If not interested in a specific labelling or extension, it is also
possible to merely prove the (non-)existence of one using automated provers for HOL.
Additionally notions of skeptical  and credulous argument justification~\cite[Defs.~56 and 57]{BCG11} have been encoded
in HOL~\cite[\texttt{base}]{Sources}, although this has not been a subject of focus here.

\section{Conclusion \label{sec:conclusion}}
In this article, an encoding of abstract argumentation frameworks as well as
various argumentation semantics into higher-order logic (HOL) was presented. To that end, 
sets are identified with their characteristic function and represented
by typed predicates within the HOL formalism. Similarly, the attack relation
of argumentation frameworks is encoded as a binary predicate of appropriate type.
Finally, argumentation semantics are represented by higher-order predicates
on extensions (labellings) that hold, by construction, if and only if 
the given extension (labelling) indeed satisfies the constraints imposed
by the respective argumentation semantics.

The presented encoding was exemplarily implemented within the
well-known proof assistant Isabelle/HOL, enabling the employment of various interactive
and automated deduction tools including, in particular, the model finder Nitpick
and the automated meta theorem prover sledgehammer.
The resulting source files of the encoding are readily and freely available at GitHub~\cite{Sources}
for further usage.
It is important to note that the encoding presented in this article is not fixed to
any one specific reasoning system; we merely chose to use Isabelle/HOL for demonstration
purposes because of its flexibility and user-friendliness, including, e.g., a graphical user interface
for interactive experimentation and its portfolio of integrated automated reasoning tools.

Due to the expressiveness of the higher-order formalism, the encoding of abstract argumentation in
HOL allows for both meta-level reasoning (i.e., reasoning \emph{about} notions of abstract argumentation)
as well as object-level reasoning (i.e., reasoning \emph{with} argumentation networks) 
using the same portfolio of first-order and higher-order automated reasoning tools. 
Both aspects were highlighted in the article in the context of Isabelle/HOL applications:
Meta-level reasoning was exemplified, firstly, by utilising Isabelle/HOL for interactively
exploring the meta-theory of abstract argumentation; secondly, the adequacy of the formalisation
itself was verified by formally encoding and proving the central properties, relationships and
correspondences from abstract argumentation literature, while obtaining counter-models for well-known non-theorems.
Subsequently, we demonstrated how to use the encoding for object-level reasoning, i.e., 
for generating extensions and labellings for given argumentation frameworks. Here, quite
specific extensions and labellings can be generated that should additionally satisfy
arbitrarily complex (higher-order) properties. Since the computation is based on a computer-verified
encoding, we automatically know that the results of the concrete outputs are correct as well.
Up to the author's knowledge, there
does not exist any other approach capable of all the above aspects.
We hence argue that an encoding of argumentation into HOL provides a uniform framework
for assessing abstract argumentation from the perspective of automated reasoning.

It has to be pointed out that the presented approach is not meant to provide an alternative
to well-established means for efficiently computing extensions (or labellings) for 
large-scale argumentation frameworks. Of course, special-purpose procedures
or SAT-based approaches do not make use of formalisms as expressive as HOL, and hence
admit decidable and, in some sense, efficient routines. In contrast, HOL automated
theorem provers are semi-decidable only. Nevertheless our aim is quite orthogonal
and rather aims at providing generic means for interactively and (partly) automatically
assess abstract argumentation within a rich ecosystem of reasoning tools. We
hence provide a bridge between the landscape of abstract argumentation one the one side
and automated deduction on the other. In fact, this is in line with the motivation 
put forward by the LogiKEy framework~\cite{DBLP:journals/ai/BenzmullerPT20}
that employs generic higher-order reasoning for assessing normative theories for
ethical and legal reasoning. As a side contribution of this work
we thus extend the LogiKEy framework with generic means of abstract argumentation,
e.g., allowing experiments in legal argumentation~\cite{DBLP:conf/itp/BenzmullerF21}
 to be based on more principled notions of argumentation.

\paragraph{Related work.}
Besides the well-known reduction to logic programs \cite{toni2011argumentation}, the
encoding of constraints enforced by argumentation semantics into other formalisms has become a standard technique for implementing abstract argumentation~\cite{HOFAimplementation}. Early works on logical encodings into propositional logic, as proposed by Dunne and Bench-Capon \cite{DBLP:journals/ai/DunneB02}, and Besnard and Doutre \cite{DBLP:conf/nmr/BesnardD04}, reduce the problem of finding admissible-based extensions as a logical satisfiability problem. This work has paved the way for later work on harnessing SAT-solvers~\cite{DBLP:series/faia/2009-185} for this task \cite{DBLP:conf/clima/WallnerWW13,HOFAimplementation}. This technique form the basis of various tools for argumentation such 
as, e.g., Cegartix~\cite{DBLP:journals/ai/DvorakJWW14}, LabSATSolver~\cite{DBLP:conf/ki/BeierleBP15} and jArgSemSAT~\cite{DBLP:journals/ijait/CeruttiVG17}.
These approaches mostly focus on generating adequate extensions (labellings) and do not
allow for introspection, i.e., support for meta-theoretical reasoning about abstract argumentation.

In a similar vein, other approaches rely on encoding abstract argumentation in more expressive logical formalisms than propositional logic. They make use of this increased expressivity for capturing many different argumentation semantics under a purely logical umbrella. An early approach towards the encoding of abstract argumentation in quantified propositional logic (QBF) has been proposed by Egly and Woltran \cite{DBLP:conf/comma/EglyW06}. An extended QBF-based approach has been introduced by Arieli and Caminada~\cite{DBLP:journals/japll/ArieliC13} to represent in an uniform way a wide range of extension-based semantics. Their approach allows for automated verification of some semantical properties, e.g., the existence of stable extensions and some inclusion and equivalence relations between extensions.
Several (restricted) first-order logical formalisms have also been proposed. Dupin de Saint-Cyr et al.~\cite{DBLP:journals/ijar/Saint-CyrBCL16} introduce a first-order language (YALLA) for the encoding of argumentation semantics and study dynamic aspects of abstract argumentation. More recently, Cayrol and Lagasquie-Schiex have proposed a first-order logical encoding of extended abstract argumentation frameworks featuring higher-order attacks and support relations~\cite{DBLP:journals/ijait/CayrolL20}.
We refer the reader to Gabbay \cite{Gabbay2013Book} for further survey and discussion about logical encodings, including modal and second-order logic-based approaches.
We note that, while those approaches are much in the same spirit as the one presented
in this article, they are less expressive and generic, since many of the meta-theoretic analyses presented in \S\ref{sec:meta-theory} cannot be carried out within them as they make essential use of higher-order constructs. This sort of expressivity limitations for existent approaches has been, of course, a conscious design choice, given the well-known expressivity vs.~scalability trade-off. In this respect, our HOL-based approach complements rather than competes with them; and it has the added value of enabling the utilisation in interactive proof assistants.

The presented encoding is closely related to the so-called \emph{shallow semantical embeddings} \cite{J41}. Such embeddings allow for the representation of domain-specific expressions by merely considering
the defined concepts as abbreviations of a host meta-language (in our case, of HOL) that can be
unfolded exhaustively, yielding an ordinary (but complex) formula in the meta-logic.
Such shallow embeddings were already studied for encoding non-classical logics into HOL,
e.g., modal logics~\cite{DBLP:journals/lu/BenzmullerP13} and many-valued logics~\cite{sixteen}. 
Via such embeddings, any higher-order reasoning system can be turned into a reasoner 
for the respective non-classical logic \cite{J41,lparModal}. This approach has previously been utilised for encoding networks of structured arguments in~\cite{DBLP:conf/clar/FuenmayorB20} and also in~\cite{DBLP:conf/itp/BenzmullerF21} in the context of legal reasoning.

\paragraph{Further work.}

We plan to further extend the collection of encoded argumentation semantics towards other recent
proposals in the literature (e.g., eager, CF2, and stage2 semantics) as well as to conduct
extended meta-theoretical studies based on them.

Moreover, the expressivity of HOL also allows us to extend the scope of our work towards other
extensions of abstract argumentation frameworks beyond Dung's approach, including
joint and higher-order attacks~\cite{HOFAadfs,DBLP:conf/birthday/BarringerGW05,baroni2011afra},
as well as bipolar argumentation that adds
support relations between arguments~\cite{DBLP:conf/ecsqaru/CayrolL05a,DBLP:journals/ijis/AmgoudCLL08}.

Given the trade-off between expressivity and scalability, we are currently exploring the limits of
our approach for larger inputs. This analysis is quite non-trivial, as it involves substantial
engineering work regarding the effective orchestration of the different automated tools in the portfolio
(automated theorem provers, SAT/SMT-solvers, model generators, etc.) for this particular family of 
applications. 

Our approach also allows, in fact, for the instantiation of abstract arguments by complex objects,
such as sets or tuples of formulas in a (non-)classical logic. This suggests a seamless extension
of our application to instantiated argumentation frameworks~\cite{HOFAinstantiated}.
Some preliminary experiments involve the instantiation of arguments as tuples composed of a pair
of formulas $\big(\texttt{support},\texttt{claim}\big)$ in a formal (recursively defined) logical language.
We can then employ the shallow semantical embedding approach~\cite{J41} to give a semantics to these
formulas, including an appropriate definition for a logical consequence relation.
Subsequently, we can then employ the latter to instantiate the corresponding attack relation between
arguments in several different ways, namely, as rebutting, undermining and undercutting, as given
by the semantics of the embedded logic. A detailed exploration of this is, however, further work.

\section*{Acknowledgements}
This work was supported by the Luxembourg National Research Fund [C20/IS/14616644 to A.S.].


\bibliographystyle{LNGAI21}
\bibliography{main}



\end{document}